\documentclass[10pt,twocolumn,letterpaper]{article}

\usepackage{wacv}
\usepackage{times}
\usepackage{graphicx}
\usepackage{amsmath}
\usepackage{amssymb}
\usepackage{mathtools}
\usepackage{bm}
\usepackage{array}
\usepackage{multirow}

\usepackage{float}
\usepackage{booktabs}
\usepackage{longtable}

\makeatletter
\@namedef{ver@everyshi.sty}{}
\makeatother
\usepackage{tikz,pgfplots}
\usetikzlibrary{calc}
\usepackage{booktabs}
\usepackage[export]{adjustbox}
\usepackage{blindtext}
\usepackage[ruled,lined]{algorithm2e}
\SetAlFnt{\small}
\SetAlCapFnt{\small}
\SetAlCapNameFnt{\small}

\usepackage{xcolor}

\newcommand{\vt}[1]{\mathbf{#1}}
\newcommand{\mat}[1]{#1}
\newcommand{\bmr}{\bm{\rho}}

\newcommand{\M}[2]{\mathcal{M}_{#1}(#2)}
\newcommand{\X}{\mathcal{X}}
\newcommand{\bbR}{\mathbb{R}}
\newcommand{\surf}[1]{\mathcal{#1}} 
\newcommand{\norm}[1]{\left\| #1 \right\|}
\newcommand{\dotp}[1]{\langle #1 \rangle}

\DeclareMathOperator{\SH}{SH}
\DeclareMathOperator{\PLS}{PLS}

\DeclareMathOperator*{\argmin}{arg\,min}

\definecolor{mlblue}{rgb}{0 0.4470 0.7410}
\definecolor{mlgreen}{rgb}{0.4660 0.6740 0.1880}
\definecolor{mlred}{rgb}{0.6350 0.0780 0.1840}

\def\wacvPaperID{1332} 

\wacvalgorithmstrack   

\wacvfinalcopy 

\ifwacvfinal
\usepackage[breaklinks=true,bookmarks=false]{hyperref}
\else
\usepackage[pagebackref=true,breaklinks=true,colorlinks,bookmarks=false]{hyperref}
\fi

\begin{document}

\title{High-Quality RGB-D Reconstruction 
via Multi-View\\ Uncalibrated Photometric Stereo 
and Gradient-SDF}

\author{Lu Sang\textsuperscript{1,2} \and  Bj\"{o}rn H\"{a}fner\textsuperscript{1,2} \and Xingxing Zuo\textsuperscript{1} \and Daniel Cremers\textsuperscript{1,2}\\
\textsuperscript{1}Technical University of Munich\\
\textsuperscript{2}Munich Center for Machine Learning\\
{\tt\small \{lu.sang, bjoern.haefner, Xingxing.Zuo, cremers\}@tum.de}
}

\maketitle
\thispagestyle{empty}

\begin{abstract}
   Fine-detailed reconstructions are in high demand in many applications. However, most of the existing RGB-D reconstruction methods rely on pre-calculated accurate camera poses to recover the detailed surface geometry, where the representation of a surface needs to be adapted when optimizing different quantities. In this paper, we present a novel multi-view RGB-D based reconstruction method that tackles camera pose, lighting, albedo, and surface normal estimation via the utilization of a gradient signed distance field (gradient-SDF). The proposed method formulates the image rendering process using specific physically-based model(s) and optimizes the surface's quantities on the actual surface using its volumetric representation, as opposed to other works which estimate surface quantities only near the actual surface. 
   To validate our method, we investigate two physically-based image formation models for natural light and point light source applications. The experimental results on synthetic and real-world datasets demonstrate that the proposed method can recover high-quality geometry of the surface more faithfully than the state-of-the-art and further improves the accuracy of estimated camera poses\footnote{source code available~\url{https://github.com/Sangluisme/PSgradientSDF}}.
\end{abstract}

\section{Introduction}

\begin{figure}[t]
   \begin{minipage}[b][][b]{0.36\linewidth}
      \includegraphics[width=0.9\linewidth]{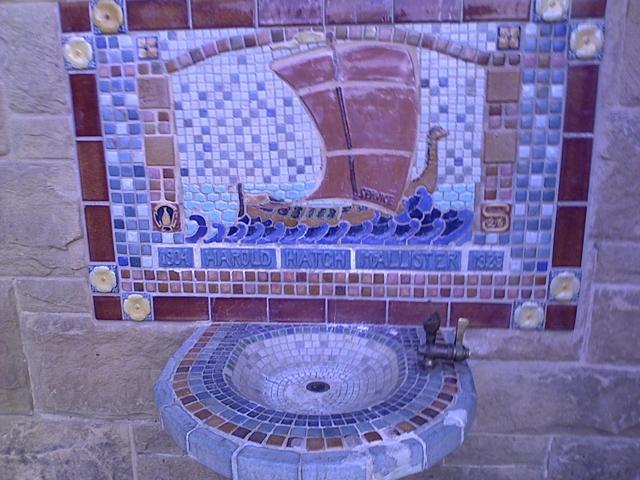} \vspace*{2mm} \\
     \centering  input RGB
   \end{minipage}%
   \begin{minipage}[b][][b]{0.36\linewidth}
      \includegraphics[width=0.9\linewidth]{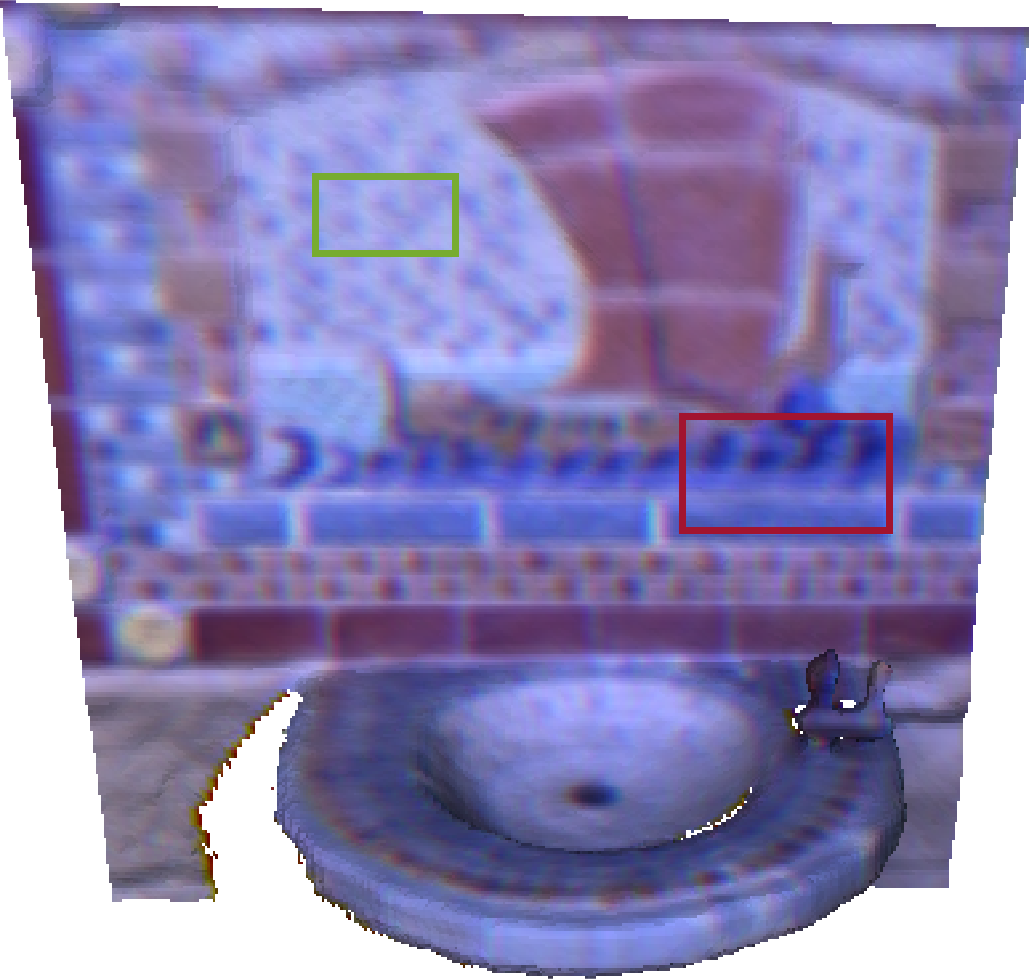} \\
      \centering \small{init reconstruction}
   \end{minipage}%
   \begin{minipage}[b][][b]{0.22\linewidth}
   \centering
   \includegraphics[width=0.9\linewidth,cframe=mlgreen 1.5pt 0pt]{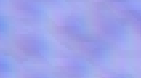}\\
   \vspace*{1mm}
   \includegraphics[width=0.9\linewidth,cframe=mlred 1.5pt 0pt]{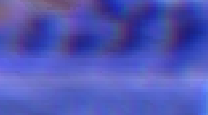} \\
   \vspace*{3mm}
   \centering init texture
   \end{minipage}

   \vspace*{1mm}
   \begin{minipage}[b][][b]{0.36\linewidth}
      \includegraphics[width=0.9\linewidth]{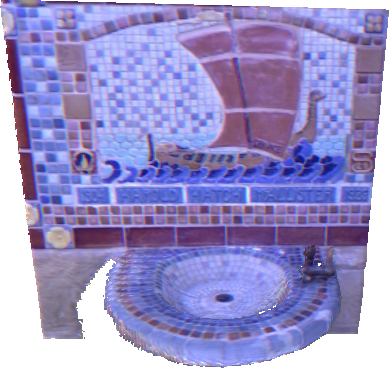} \\
     \centering  estimated albedo
   \end{minipage}%
   \begin{minipage}[b][][b]{0.36\linewidth}
      \includegraphics[width=0.9\linewidth]{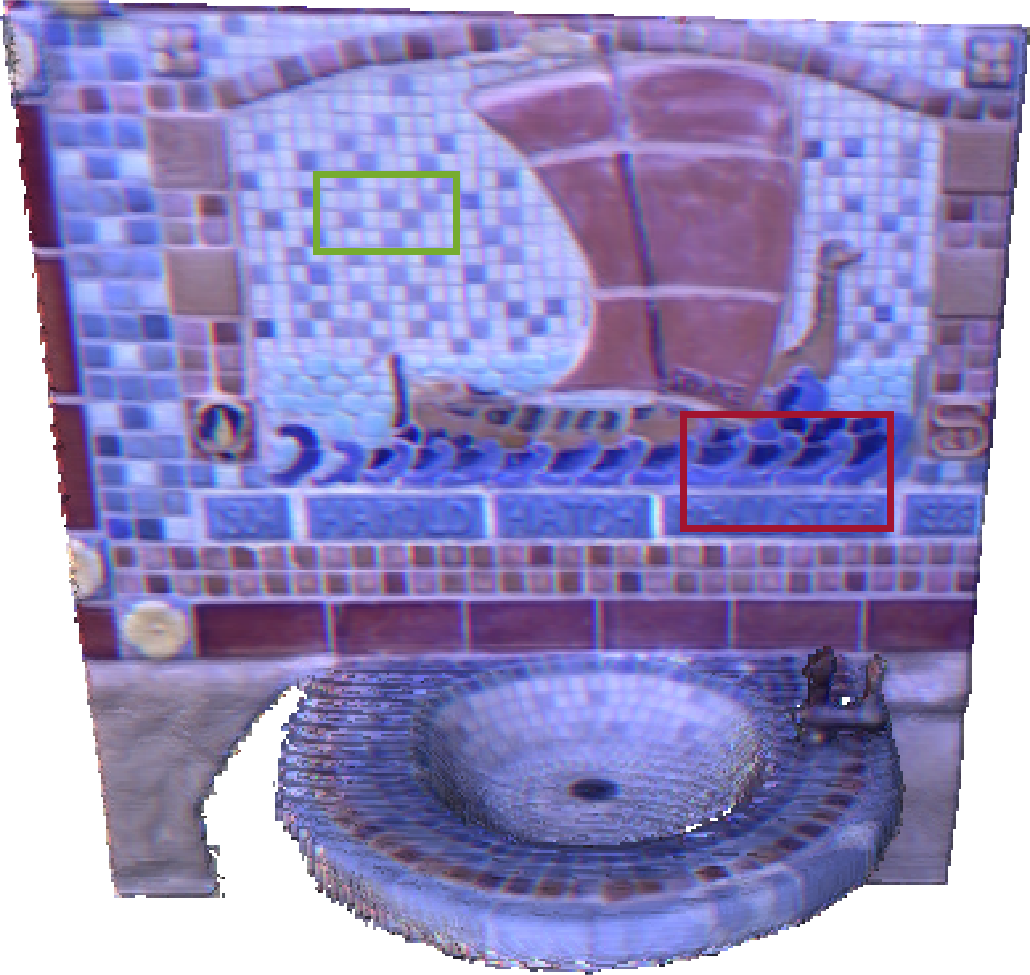} \\
      \centering \small{refined reconstruction}
   \end{minipage}%
   \begin{minipage}[b][][b]{0.22\linewidth}
   \centering
   \includegraphics[width=0.9\linewidth,cframe=mlgreen 1.5pt 0pt]{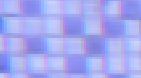}\\
   \vspace*{1mm}
   \includegraphics[width=0.9\linewidth,cframe=mlred 1.5pt 0pt]{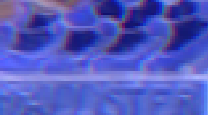} \\
   \vspace*{3mm}
   \centering \small{refined texture}
   \end{minipage}
   \vspace{1.5mm}
   \caption{First row: one example input RGB image~\cite{zhou2014color} and the initial reconstruction with zoomed-in detail texture. Second row: estimated albedo, refined reconstruction with zoom-in detail textures of the proposed method. We jointly estimate camera pose, surface normal, albedo and lighting to enable a fine-scaled 3D reconstruction.}\label{fg:tasser}
\end{figure}

 

   

Detailed surface reconstruction from 2D images and depth is a challenging topic in computer vision, with increased attention in recent years. It is required to not only reconstruct the rough shape of an object but also to recover the rich texture and fine geometric details of the surface. The reconstructed results can be used in many applications, such as 3D printing~\cite{gokhare2017review}, virtual reality~\cite{hull2007}, digital media~\cite{jorge2018}, etc. To recover a fully detailed 3D model, camera poses and surface quantities such as normal and reflectance (albedo) must be estimated. Instead of assuming known camera poses and directly optimizing surface quantities only \textit{near} the surface, in this paper, we propose to tackle the surface refinement problem together with camera poses, lighting, and albedo via utilization of gradient-SDF~\cite{Sommer2022gradient}, a volumetric representation, allowing us to \textit{optimize surface quantities directly on the surface}. In summary, we propose the following main contributions:
\begin{itemize}
   \item A novel formulation of physically realistic image model(s) compatible with a volumetric representation which enables effective optimization on actual surface points.
   \item A complete reconstruction pipeline with the capability of camera pose, geometry refinement, albedo and lighting estimation.
   \item  A method that enables camera pose refinement while taking the lighting condition into account.
   \item A method that deals with natural and point light source scenarios in an uncalibrated setting which are convenient to adopt in real applications.
\end{itemize}
Moreover, extensive examination of our approach on both synthetic and real-world datasets demonstrates that it can perform high-quality reconstruction of fine-scaled geometry, albedo, lighting, and camera tracking. 

\section{Background and Related work}\label{se:related_work}
Two critical problems must be considered when recovering a 3D model: the choice of surface representation and the underlying image formation model together with its inverse rendering techniques.\par
A surface is a $2$D manifold and the surface representation strategy is how to discretize, store and update the continuous surface. It can mainly be divided into two categories, explicit representation and implicit representation. The explicit representation, such as point clouds~\cite{Soltani2017}, surfels~\cite{schops2019bad}, or meshes~\cite{lorensen1987marching}, store the exact surface point locations, allowing for operations on the surface point itself. The implicit representation, such as signed distance function (SDFs)~\cite{Osher2003}, volume density~\cite{mildenhall2020nerf} or occupancy~\cite{peng2020convolution}, on the other hand, only store related properties of each unit, e.g., the distance to the surface~\cite{newcombe2011kinectfusion}. Different representations are appreciated by different goals. For instance, camera tracking and refining, i.e., bundle adjustment, as well as image rendering models benefit from the explicit representation because the models are built on the actual surface. Geometry-related reconstruction methods, such as KinectFusion~\cite{newcombe2011kinectfusion} prefer implicit representation because a uniform and smooth surface can be extracted from it, while easily allowing topological changes during optimization. Therefore, many works alternate between two presentations to deal with different parameters such as camera pose and surface quantities~\cite{niessner2013real,lee2020texturefusion}.\par
For the surface modeling, nowadays, there are different methods for recovering a rough 3D model, such as depth fusion-based methods~\cite{newcombe2011kinectfusion} or RGB-image-based structure-from-motion technique~\cite{schoenberger2016sfm}. However, the reconstructed 3D model lacks desired geometry detail. To further recover the fine-scale geometry, several strategies are applied. For example, by improving the accuracy of the input, e.g., depth quality~\cite{sang2020wacv, he2021towards} or camera poses accuracy~\cite{schops2019bad, niessner2013real}. Alternatively, by increasing the resolution of the chosen surface representation, e.g., the work from Lee el at.~\cite{lee2020texturefusion} introduces a texture map attached to each voxel and subdivides the texture map for a higher texture representation, but the geometry resolution is not improved. Nevertheless, these methods do not use the underlying physical relation between the RGB images, lighting condition, and the surface geometry. \par

\begin{figure}[ht]
   \begin{tabular}{cc}
      \includegraphics[width=0.45\linewidth]{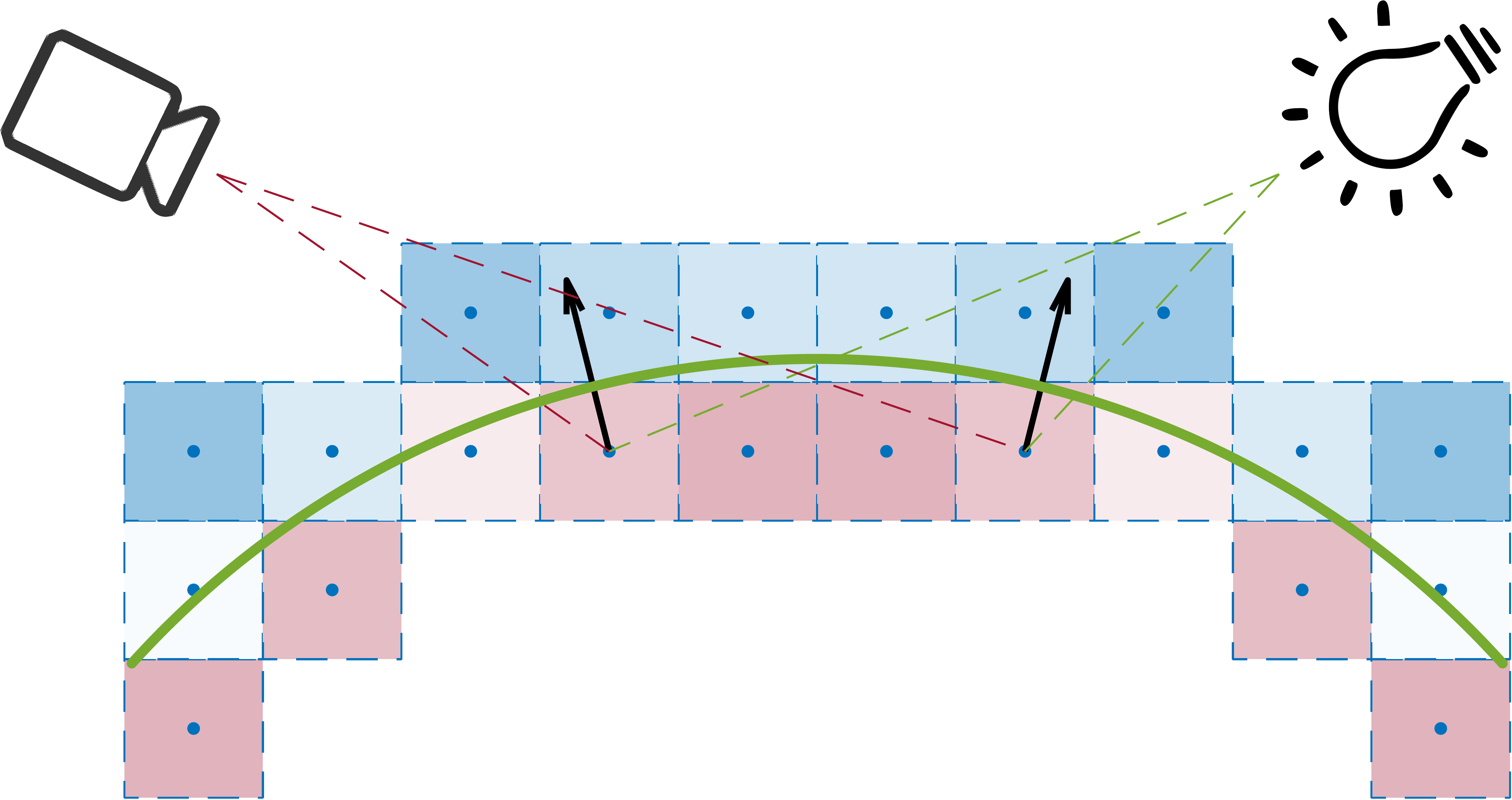} &
      \includegraphics[width=0.45\linewidth]{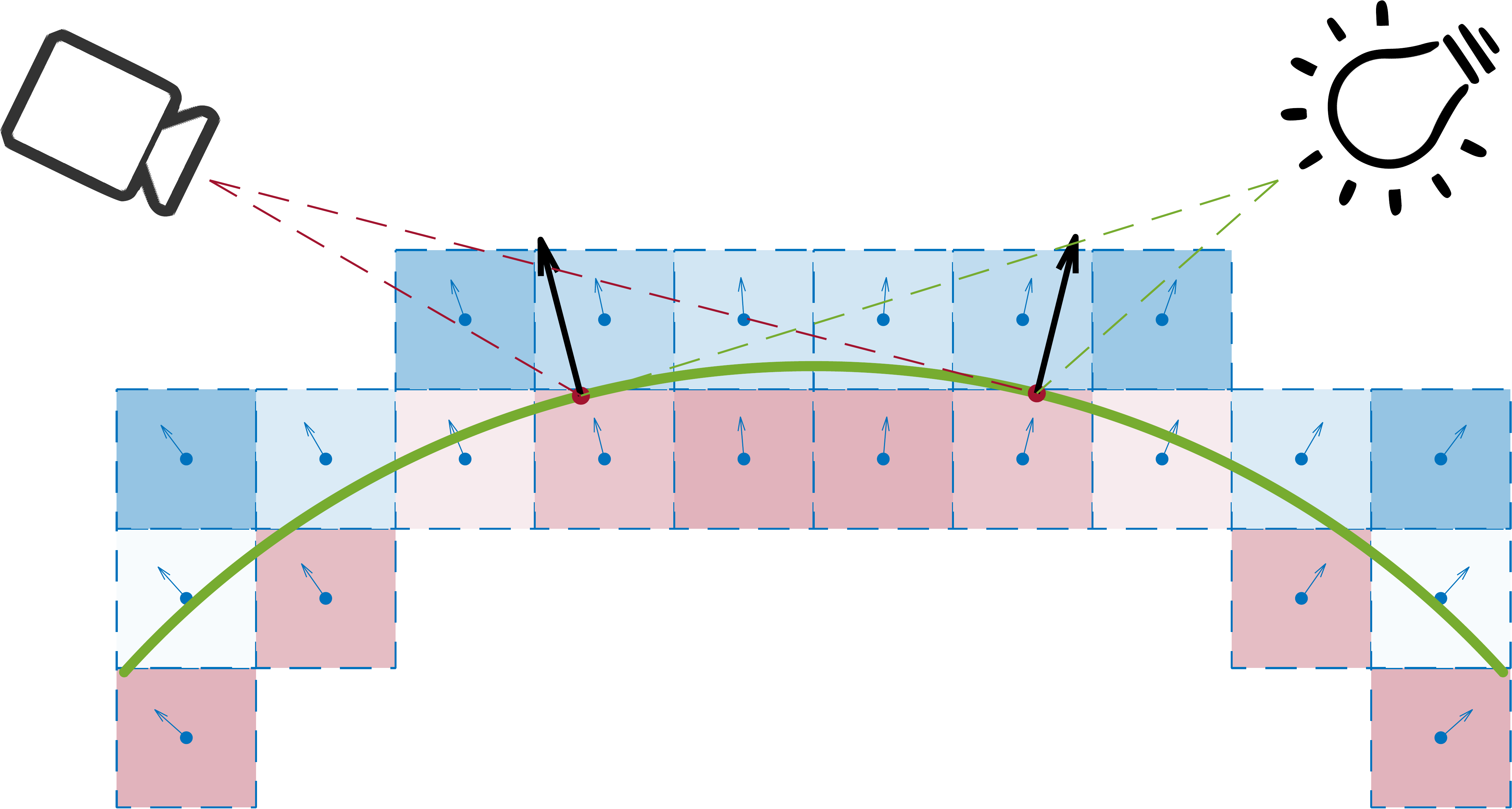} \\
      voxel center & surface point
   \end{tabular}
   \caption{Compared to modeling the surface irradiance on the voxel center in a volumetric surface representation (left), the proposed method models the scene more precisely on the actual surface (right) by moving from the voxel center along the surface normal direction.}
   \label{fg:voxel_surface}
\end{figure}

In recent years, \emph{Photometric Stereo} (PS)~\cite{Woodham1977Reflectance} methods are widely applied in different research fields, such as geometry recovery~\cite{maier2017intrinsic3d, bylow2019combining,sang2020wacv} and image rendering~\cite{yu2021plenoxels, physg2020}. A PS model describes the physical imaging process, which reflects the irradiance and the outgoing radiance of surfaces. The irradiance is affected by lighting conditions, while the radiance typically depends on the surface material and normal. Therefore, using the image formation model to formulate the rendering equation, the desired surface properties such as surface normal, texture, and material can be 
recovered~\cite{basri2001}. However, using one single RGB image to restore the desired quantities is an ill-posted problem~\cite{brahimi2020springer}. To overcome the ill-posedness, one can use several RGB images as input~\cite{haefner2019iccv,maier2017intrinsic3d, sang2020wacv}. Those images can be at the same view point~\cite{peng2017}, but to recover a complete 3D model, one needs the images from different camera positions.
Existing algorithms that use PS and RGB-D sequences to recover a full 3D model~\cite{bylow2019combining, maier2017intrinsic3d, zollhofer2015shading} need pre-calculated camera poses. These works follow the pipeline that first integrates the depth with known camera poses to a volumetric cube, then re-projects each voxel back to the images and minimizes the PS energy. They archive good results; still, there are two drawbacks. First, as Figure~\ref{fg:voxel_surface} shows, they actually evaluate all optimization quantities at the voxel center instead of the surface point. Even with smaller voxel sizes, this gap might be reduced, but smaller voxel sizes limit the 3D model's size or introduce heavy computational overhead when the grid size needs to be enlarged. Most works use different regularizers to reduce the artifices caused by this gap. For instance, re-projecting a voxel to the depth map
~\cite{bylow2019combining, maier2017intrinsic3d, zollhofer2015shading} to constraint the voxel distance updating, or constrainting the total variations of reflectance~\cite{bylow2019combining, maier2017intrinsic3d}. The second drawback is that most of the works rely on an independent camera tracking algorithm for real-world datasets to get the initialized camera poses. These tracking algorithms either use pure depth information~\cite{newcombe2011kinectfusion} or assume consistent light conditions across the input RGB images~\cite{niessner2013real, bylow2013real,schoenberger2016sfm}. Thus, they do not employ color information, or their assumption contradicts the image formation model.  

Notably, there is another class of methods: neural rendering. The success of the neural rendering methods leads to a boom of learning-based methods for image-based synthesis and geometry recovery. Some neural rendering methods also employ PS methods, but most of these approaches~\cite{yu2021plenoctrees, physg2020} focus on view synthesis rather than geometry refinement, and their inputs are only RGB images with accurate camera poses. The work of Lin et al.~\cite{lin2021barf} enables camera pose refinement, still it does not focus on recovering geometry details or estimating the reflectance and lighting. The recent RGB-D neural rendering work~\cite{azinovic2021neural} does not apply to most real-world datasets since it assumes single focal length camera model. To the best of our knowledge, only these two papers~\cite{Yariv2021, wang2021neus} concentrate on the recovery of the geometry and filling the gap between voxel representation and surface points representation in a slightly different way. They propose to transfer the signed distance value to the density of the point and then use the accumulated color along the ray to recover a good 3D geometry, hence the surface representation as points and volume are combined. Yet, they learn each frame independently rather than explicitly model the lighting and surface normal as PS methods, Even so, they do not refine camera poses and have a long training time. \par
In this paper, we propose a method that tackles the fine-detailed 3D reconstruction problem together with camera pose estimation in a complete pipeline that stays in one consistent surface representation and assumption. Moreover, we estimate the surface quantities at the actual surface and jointly optimize camera pose after initializing the camera pose and coarse surface volume using gradient-SDF~\cite{Sommer2022gradient}.
\section{Image Formation Model}\label{se:method}
\subsection{Surface Reflectance}\label{subse:two_models}
Photometric stereo techniques study the reflected light from the surface together with the environment lighting, the surface material, and normal based on the physical laws to model the produced color images. It describes the outgoing radiance of a surface point $\vt{x} \in \mathbb{R}^3$ by integrating over the upper hemisphere $\mathbb{S}^2_+$ around the surface normal $\vt{n}\in\mathbb{S}^2$. The pixel intensity that is conjugated with this point on the image is proportional to its irradiance. The outgoing direction can be regarded as the camera view direction. Therefore, for an image $\vt{I}$, the color of the pixel $\vt{p}\in\mathbb{R}^2$ conjugated with $3$D point $\vt{x}\in\mathbb{R}^3$ is
\begin{equation}
   \label{eq:image_formation}
       \vt{I}(\vt{p}(\vt{x})) \approx \int_{\mathbb{S}^2_+} \bmr(\vt{x},\vt{i}) \vt{L}(\vt{i},\vt{x})\max(\dotp{\vt{i},\vt{n}(\vt{x})}, 0) \mathrm{d}\vt{i}\,,
\end{equation}
where $\bmr:\mathbb{R}^3\times\mathbb{S}^2\times\mathbb{S}^2\to\mathbb{R}^3$ is the bidirectional reflectance distribution (BRDF) function with $3$ color channels, 
and $\vt{L}:\mathbb{S}^2\times\mathbb{R}^3\to\mathbb{R}^3$ 
is the incoming lighting radiance from direction $\vt{i}\in\mathbb{S}^2$ at point $\vt{x}\in\mathbb{R}^3$~\cite{Kajiya1986}. Assuming a constant BRDF, i.e., a Lambertian surface,~\eqref{eq:image_formation} can be simplified as 
\begin{equation}
   \label{eq:diffuse_formation}
       \vt{I}(\vt{p}(\vt{x})) = \bmr(\vt{x}) \int_{\mathbb{S}^2_+}  \vt{L}(\vt{i},\vt{x})\max(\dotp{\vt{i}, \vt{n}(\vt{x})}, 0)\mathrm{d}\vt{i}\,.
\end{equation}
Given enough color images, the surface normal $\vt{n}$ and the surface reflectance $\bmr$ can be therefore restored. The crucial challenge is to approximate the integral over the hemisphere reasonably. Different models are proposed to simplify the integral to solve the equation. Here we introduce two models that deal with two application scenarios: natural lighting and point light source illumination.
\paragraph*{Natural Light Spherical Harmonics Model}
A \emph{natural light} source situation is, for example, the light source is the sun, or when the light sources are in a distance. The lighting directions that reach the surface are nearly parallel, then the environment lighting can be well modeled using Spherical Harmonics (SH) functions. The integral part is approximated by the sum of SH basis~\cite{basri2001, basri2003}.
\begin{equation}
\label{eq:SH_model}
   \vt{I}(\vt{p}(\vt{x})) \approx \bmr(\vt{x})\dotp{\vt{l}, \SH(\vt{n}(\vt{x}))}\,,
\end{equation}
where $\vt{l} \in \bbR^4$ is an $4$-dimensional lighting vector for the current view and $\SH(\vt{n}(\vt{x}))\in \bbR^4$ are the first-order SH basis functions for a fixed $\vt{n}(\vt{x})$. 
The model is simply formulated and still can reach relatively high accuracy with a low order model. 

\paragraph*{Point Light Source Model}
Apart from the natural light, another commonly encountered scenario is the point light source situation, mainly when focusing on small object reconstruction. The object is usually illuminated by a \emph{point light source}, e.g., a LED light that is close to the object. The lighting can hardly be regarded as a set of parallel lines, thus point light source provides more changes to the object illumination, which is preferred to deal with the ill-posedness of the PS model. One widely used point light source-light model is~\cite{Logothetis2018, Mecca2014, Queau2017}.
\begin{equation}
    \label{eq:LED_model}
    \vt{I}(\vt{p}(\vt{x})) = \Psi^s \bmr(\vt{x}) (\frac{\dotp{\vt{n}^s,\vt{l}^s}}{\norm{\vt{l}^s}} )^{\mu^s} \frac{\max(\dotp{\vt{n}(\vt{x}),\vt{l}^s}, 0)}{\norm{\vt{l}^s}^3}\,,
\end{equation}
where $\Psi^s$ is the light source intensity, $\vt{n}^s$ is the principal direction of the light source, and $\vt{l}^{s}$ is the vector pointing from the light location to the surface point. The denominator term $\norm{\vt{l}^{s}}^3$ describes the attenuation of the light intensity when it reaches the surface point. $\mu^s \geq 0$ is the anisotropy parameter. Typically, when using point light source model, other than the fact that the model is highly non-linear and non-convex, which makes it hard to optimize. Another challenge is that many parameters need to know or optimize, such as light source location and principal direction. Most works with point light source model have additional steps to calibrate the light sources intensity, location, and principal direction~\cite{logothetis2019differential, Queau2017}. The work of Logothetis el at.~\cite{logothetis2019differential} designs a device with LED lights arranged circularly on a board with a centered camera and switches on and off the LED lights to capture the images in different lighting. Then they refine a pre-generated SDF model. The work of Qu\'{e}au et al.~\cite{Queau2017} only takes RGB images as input and jointly optimizes the depth and reflectance. However, their output is a single depth image and needs strictly calibrated camera poses and light sources. In the next section, we will explain how our proposed model works without tedious lighting calibration, thus making our approach completely uncalibrated in the point light source and natural lighting setting.

\subsection{Multi-view PS Models}
\begin{figure}[b]
   \centering
   \includegraphics[width=0.7\linewidth]{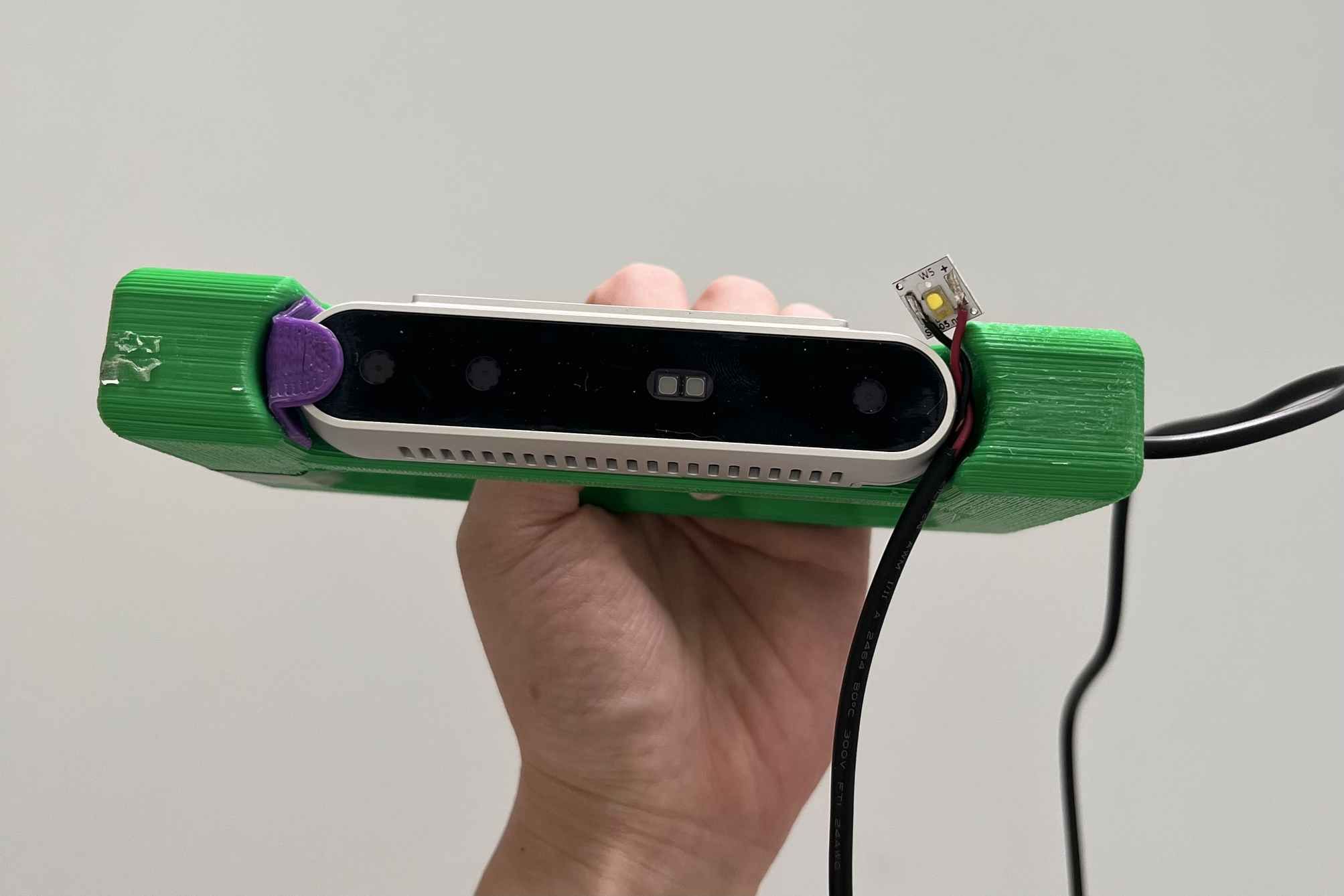}
   \caption{The simple set-up we used to record the datasets, an Intel Realsense D415 camera with an LED light attached to it.}\label{fg:set_up}
   \vspace*{1.5mm}
\end{figure}

To recover the complete $3$D model, a sequence of RGB images $\{\vt{I}_i\}_i$ for $i\in\mathcal{I}$ are given with camera-to-world pose pairs $\{\mat{R}_i, \vt{t}_i\}_i$. 
Denote the $3$D point $\vt{x}$ warped to image $\vt{I}_i$ as $\vt{x}_i = \mat{R}_i^\top\vt{x}-\vt{t}_i$, and plugging this into~\eqref{eq:SH_model} and~\eqref{eq:LED_model}, we write the multi-view models residual as 
\begin{equation}
   \vt{r}_{i} = \vt{I}_i(\vt{p}(\vt{x}_{i})) -  \bmr(\vt{x})\mathcal{M}(\vt{x}, \mathcal{X}_i)\,,
   \label{eq:5}
\end{equation}
$\M{\cdot}{\cdot, \cdot}$ stands for two different image formation models and $\mathcal{X}_i$ is the variables in the two image formation models introduced in Section~\ref{subse:two_models}. Hence, the SH model is
   \begin{equation}
      \M{\SH}{\vt{x}, \mathcal{X}_i} = \dotp{\vt{l}_i, \SH(\mat{R}_i^\top\vt{n}(\vt{x}))}\,.
   \label{eq::SH_model_multi_view}
\end{equation}
Point $\vt{x}$, as well as its normal is transferred to the image coordinates by camera pose $\mat{R}_i$ and $\vt{t}_i$. Here $\X_i = (\mat{R}_i, \vt{t}_i, \vt{l}_i)$. \\
For the point light source model, we propose a setup similar to~\cite{sang2020wacv}, i.e. attaching an LED light to the camera when capturing the images, see Figure~\ref{fg:set_up}.
Since the camera poses are known, the light position in~\eqref{eq:LED_model} is at the origin of the camera coordinates, and the light source principal direction can be regarded as the $-z$ direction, i.e., we assume we have a collocated light-camera setup for our point light source model. The $\vt{l}^s$ in Equation~\eqref{eq:LED_model} is then pointing from origin to the surface point $-\vt{x}_i$ under the camera coordinates. If additionally, we assume an isotropic light source, i.e., $\mu^s = 0$, the point light source model with multi-view setting reads
\begin{equation}
   \label{eq::LED_model_multi_view}
   \M{\PLS}{\vt{x}, \mathcal{X}_i} = \Psi^s_i \bmr(\vt{x}) \frac{\max(\dotp{\mat{R}_i^\top\vt{n}(\vt{x}),-\vt{x}_{i}}, 0)}{\norm{\vt{x}_{i}}^3}\,,
\end{equation}
where $\X_i = (\mat{R}_i, \vt{t}_i, \Psi^s_i)$. 

\subsection{Geometry and Camera Poses Initialization}\label{subsec:tracking}
Geometric and photometric error are the two error terms that are typically utilized in camera tracking techniques. While the photometric error evaluates the color constancy when projecting one point to another RGB image, the geometric error determines the depth displacement when warping points to another depth frame. The majority of earlier surface refinement techniques~\cite{lee2020texturefusion,ha2021normalfusion,maier2017intrinsic3d,bylow2019combining} use tracking methods such as~\cite{schoenberger2016mvs} or~\cite{niessner2013real,bylow2013real}, where photometric error terms are adapted. Therefore, colors are assumed to be consistent across images when initializing the camera poses, but are assumed to be different later in the image formation model. To avoid this inconsistency, the proposed method initializes the SDF volume and estimated camera pose only using depth information, then optimize the camera poses together with surface properties taking lighting conditions into account later. The camera pose for $i$-th frame is optimized by aligning the point cloud with points $\vt{x}^k$ from depth $i$ to the global shape $\surf{S}$ by~\cite{bylow2013real, Sommer2022gradient}
\begin{equation}
   \label{eq:tracking}
   \min_{\mat{R}_i, \vt{t}_i} E(\mat{R}_i,\vt{t}_i) = \sum_{k}{w_i^kd_\surf{S}(\mat{R}_i\vt{x}^k+\vt{t}_i)^2}\,,
\end{equation}
where 
$w^k_i = \max(\min(1+\frac{d_i^k}{T}, 1), 0)$ is the truncated SDF weight of points $k$ at frame $i$, $T$ is the truncated distance and $d_\surf{S}(\vt{x})$ is the distance from point $\vt{x}$ to the shape $\surf{S}$. 
Please refer~\cite{bylow2013real} or supplementary material for more details.

\section{Voxel Based Photometric Modeling} \label{se:modeling}
\subsection{From Voxel to Surface}
The key idea of the proposed method is describing the explicit surface under an implicit representation, i.e., under a volumetric cube. To perform all the operations on the actual surface, we must find the corresponding surface point to each voxel. To combine the advantage of surface point representation with volumetric representation, gradient-SDF~\cite{Sommer2022gradient} stores the signed distance $\psi^j$ for each voxel $\vt{v}^j$ for $j\in\mathcal{V}$, together with the distance gradient $\vt{g}^j$ of this voxel. It allows us to easily compute the surface point $\vt{x}^j$ by moving along the gradient direction $\vt{g}^j$ with the voxel distance $\psi^j$, 
\begin{equation}
   \label{eq:surface_point}
    \vt{x}^j = \vt{v}^j - \vt{g}^j\psi^j\,.
\end{equation}
We then can develop our model precisely on the surface points. We will show it is not only theoretically more precise, but also leads to better quantitative results in Section~\ref{se:evaluation}. To retrieve the color of a surface point in the image, we can project the actual surface back to the image domain by
\begin{equation}
    \vt{I}_i(\vt{p}(\vt{x}_i^j)) = \vt{I}_i(\pi(\mat{R}_i^\top(\vt{v}^j-\psi^j\vt{g}^j-\vt{t}_i))) \,,\label{eq:surface_on_image}
\end{equation}
where $\pi: \mathbb{R}^3 \to \mathbb{R}^2$ is the projection operator that maps 3D points to 2D pixels on the RGB image and $\vt{x}_i^j$ is $j$-th point projected on $i$-th image. The stored voxel gradient $\vt{g}$ is the surface normal $\vt{n}(\vt{x})$ as described in~\cite{Sommer2022gradient}, and $\vt{I}_i$ is the $i$-th color image. 
To recover the reflectance field (albedo) of the surface, for each voxel we directly save the estimated albedo of the surface point, i.e., $ \bmr^j = \bmr(\vt{v}^j - \psi^j\vt{g}^j)$.


\subsection{Multi-view PS Energy}
We now present the multi-view volumetric-based photometric stereo model. For a set of input RGB images $\{I_i\}_i$ and a set of gradient-SDF voxels $\{\vt{v}^j\}_j$, the following energy function is minimized to recover the high-quality texture and jointly perform the bundle adjustment for camera pose refinement.
 \begin{align}  \label{eq:energy}
    \min_{\{\bmr^j, \psi^j\}_j, \{\X_i\}_i}&\vt{E}(\bmr^j, \psi^j, \X_i) \nonumber \\ 
   &= \sum_{i\in\mathcal{I},j \in \mathcal{V}} \nu_i^j \Phi \big( \vt{I}_{i}^j - \bmr^j\M{}{\vt{x}^j, \X_i} \big) \nonumber \\ 
    &+ \lambda \sum_{j \in \mathcal{V}} | \norm{\nabla \psi^j}^2 - 1|^2 \,, 
\end{align}
where $\Phi(\cdot)$ is a robust M-estimator~\cite{Queau2017}. We choose Cauchy estimator $\Psi(x) = \log(1+ \frac{x^2}{\sigma^2})$. The parameters $\sigma = 0.2$. $\nu_i^j$ is the visibility map of the $j$-th voxel in image $i$ which we stored during camera tracking stage. 
$\vt{I}_i^j = \vt{I}_i (\vt{p}(\vt{x}_i^{j}))$ as described in equation~\eqref{eq:surface_on_image}. $\M{}{\vt{x}^j, \X_i}$ is as stated in~\eqref{eq::LED_model_multi_view} and~\eqref{eq::SH_model_multi_view} with point $\vt{x}^j$ is adapted from voxel $\vt{v}^j$ using~\eqref{eq:surface_point}. For each voxel two variables $\bmr^j$ and $\psi^j$ are optimized. Both models have camera poses embedded; therefore, the camera poses can also be improved by minimizing the model. Joint optimization of camera poses and surface geometry, albedo, and lighting is a superior choice and leads to better results, as we will show in Section~\ref{se:evaluation}. \par

Note that we only need one regularizer in~\eqref{eq:energy} on the distance gradient to ensure the distance satisfies the Eikonal equation. The regularizer needs to be made aware of two important things. First, because there is no voxel-center-to-surface gap, our energy function is straightforward and elegant. Contrary to the previous works~\cite{bylow2019combining, maier2017intrinsic3d}, our formulation~\eqref{eq:energy} is not empirical but has a physical meaning: for 2D manifolds embedded in 3D being represented with an SDF, the signed distance field is differentiable and its gradient satisfies the Eikonal equation~\cite{Osher2003} on the iso-surface. The regularizer guarantees that the updated distance is still within the distance field. 
 Second, constraining only the distance field itself detaches the energy from depth images. We do not need to store depth images during image formation model optimization of~\eqref{eq:energy}. \par

Note that the geometry is refined by optimizing the normal as well as the distance of a gradient-SDF voxel. Using the fact that close to the iso-surface, the gradient of the distance and surface normal satisfy
\begin{equation}\label{eq:normal}
\vt{g}^j = \frac{\nabla \psi^j}{\norm{\nabla \psi^j}} \,.
\end{equation}

\subsection{Voxel Up-sampling}
A smaller voxel size is preferable for representing fine-scaled geometry details. 
However, for the previous SDF-based method~\cite{maier2017intrinsic3d}, the up-sampling is possible by interpolating between neighborhood voxels. Several voxels need to be accessed for one subdivision. Our proposed method can effectively up-sample and only one voxel is needed for up-sampling $2\times 2\times 2=8$ sub-voxels to reach $(\frac{v^s}{2})$ voxel size using Taylor expansion similar as~\eqref{eq:surface_point}
\begin{align}
   \label{eq:taylor}
    \vt{v}^j_{1\dots 8} &= \vt{v}^j + \frac{v^s}{4}\vt{s}_{1\dots 8} \,, \\
    \Rightarrow \quad
    \psi^j_{1\dots 8} &= \psi^j + \frac{v^s}{4}\left(\vt{s}_{1\dots 8}\right)^\top\vt{g}^j\,,
\end{align}
where $\vt{s}^{1\dots 8} =(\pm 1, \pm 1, \pm 1)^\top$ indicate the $8$ different directions from the coarse voxel. The distances of the sub-voxels are therefore re-initialized using the coarse voxel. The gradient is set to be the same of the coarse voxels because it will be updated in the next optimization steps. 
We include the up-sample around surface feature in our algorithm to enable higher resolution reconstructions.

\subsection{Optimization Pipeline}
We alternatingly update the surface quantities and camera poses during optimization. 
The $\{\bmr^j\}_j$ is initialized using the average intensities of voxel $j$. The optimization pipeline is as shown in Algorithm~\ref{al:voxelps}.
\begin{algorithm}[ht]
   \LinesNotNumbered
   \SetKwInOut{Input}{input}
   \SetKwFor{For}{for (}{) }{}
   \SetAlgoLined
     \Input{$\{(\psi^j,\bmr^j)\}_j$, $\{\mat{R}_i, \vt{t}_i\}_i$, $\{I_i\}_i$}
     \BlankLine
     \While{$k<$ max iter and not converge}{
      \If {up-sample}{
         $\mathcal{V} \leftarrow \mathcal{V}_{up}, \psi^j \leftarrow \psi^j_{up}$
      }
      \For{$j \in \mathcal{V}$}{
     $\bmr^j \leftarrow \min\vt{E}(\psi^{j,(k)},\bmr,\mat{R}_i^{(k)}, \vt{t}_i^{(k)},\vt{l}_i^{(k)})$,}
     \For{$i \in \mathcal{I}$}{
     $\vt{l}_i \leftarrow \min\vt{E}(\psi^{j,(k)},\bmr^{j,(k+1)},\mat{R}_i^{(k)}, \vt{t}_i^{(k)}, \vt{l}_i)$,}
     \For{$j \in \mathcal{V}$}{
     $\psi^j \leftarrow \min\vt{E}(\psi,\bmr^{j,(k+1)},\mat{R}_i^{(k)}, \vt{t}_i^{(k)},\vt{l}_i^{(k+1)})$, \\
     $\hat{\vt{g}}^j \leftarrow \nabla \psi^j$,}
     \For{$i \in \mathcal{I}$}{
     $\mat{R},\vt{t}_i \leftarrow \min\vt{E}(\{\bmr, \psi\}^{j,(k+1)}, \mat{R}_i, \vt{t}_i, \vt{l}_i^{(k+1)})$,}
     }  
   \caption{Optimization Pipeline}\label{al:voxelps}
\end{algorithm} 
The convergence condition is satisfied when the relative difference of $k$-th step energy and $(k-1)$-th step energy is smaller than the convergence threshold. 

\section{Evaluation}\label{se:evaluation}
To demonstrate the results of the proposed method, due to the fact that it lacks method that offers both camera pose and surface geometry refining estimation simultaneously, we divide our evaluation into two parts. The first part is a quantitative evaluation of the refined camera poses. We test our method on the TUM RGB-D~\cite{sturm2012benchmark} benchmark where the ground truth camera poses are provided. We compare the refined camera pose with three other state-of-the-art tracking methods~\cite{bylow2013real, schops2019bad, Sommer2022gradient}. The second part is the evaluation of surface refinement. We compared against four state-of-the-art methods; two classical approaches~\cite{maier2017intrinsic3d, bylow2019combining} and two learning based approaches~\cite{wang2021neus, Yariv2021} on both synthetic and real world datasets. 
\paragraph*{Set-up and Run Time} Our data structure is implemented in C++ with single-precision \texttt{float}s
. All experiments are performed on an Intel Xeon CPU @ $3.60\,\text{GHz}$ without GPU. For both surface quantities optimization and pose refinement, we use the damped Gauss-Newton method~\cite{bjoern1996} with $\lambda = 0.1$. The convergence threshold is $10^{-3}$. We only enable the up-sampling once after $5$ iterations. More mathematical and experiments details are stated in the supplementary material. We use an LED light attached to an Intel Realsense D$415$ RGB-D camera with a hand-held bar, see Figure~\ref{fg:set_up} to record the point light source datasets. For synthetic datasets we use $256^3$ voxel grid with $2$cm voxel size, resulting in an initial
 SDF of $512$MB, after integrating the depth maps. We only store voxels that contain surface points for optimization (for the synthetic bunny dataset~\cite{bunny}, it is around $20$k points). The camera tracking needs around $300$ms per frame, and each optimization iteration needs around $8$s per variable. The method converges typically after $\sim\!\!20$ iterations.

\subsection{Camera Poses Refinement}
\begin{figure}[!h]   
   \begin{minipage}{0.75\linewidth}
      \centering
      \includegraphics[width=\linewidth]{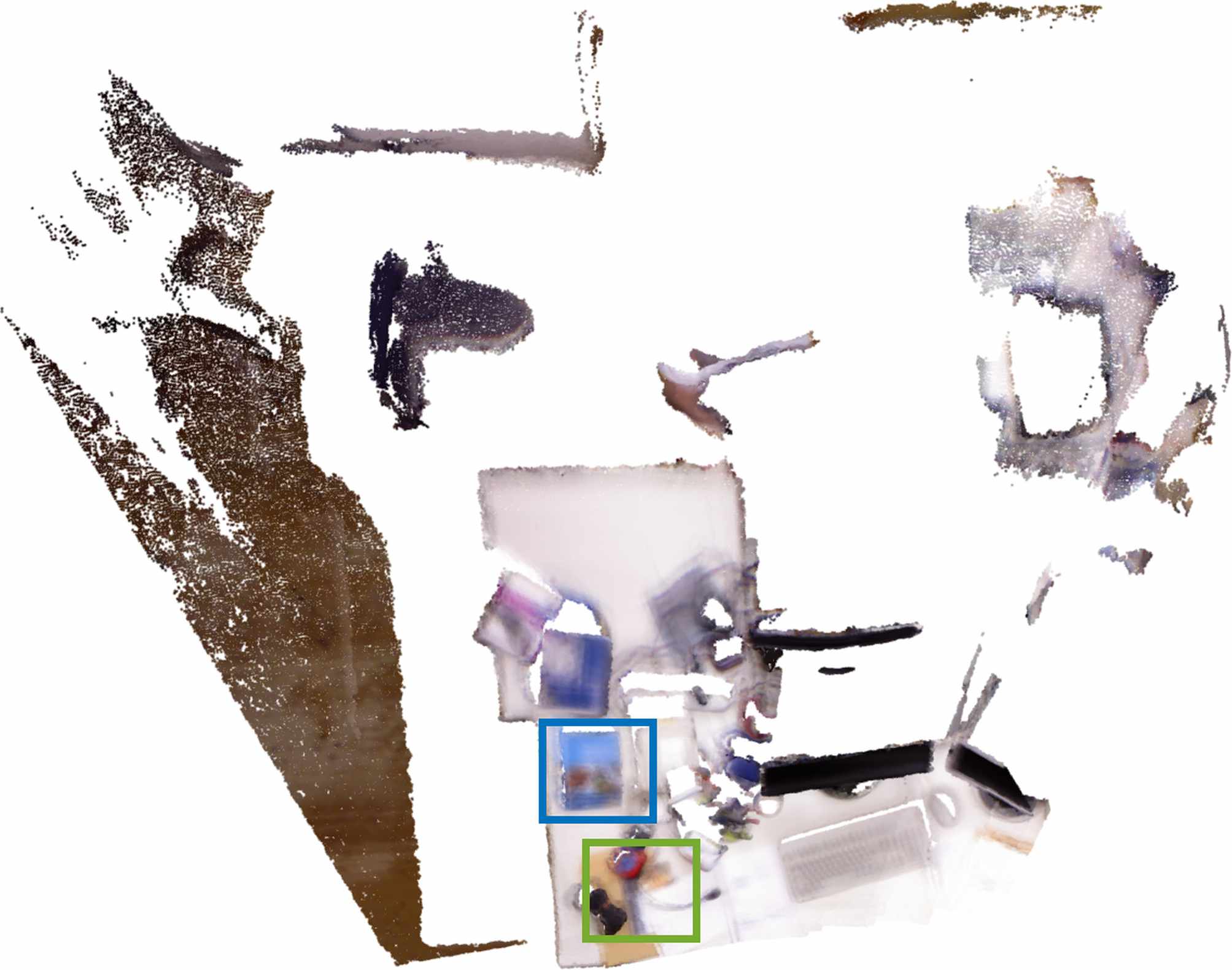}
   \end{minipage}%
   \begin{minipage}{0.2\linewidth}
   \centering
   \includegraphics[width=0.9\linewidth,cframe=mlgreen 1.5pt 0pt]{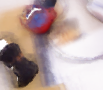}
   \includegraphics[width=0.9\linewidth,cframe=mlblue 1.5pt 0pt]{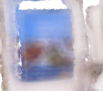}
   \end{minipage}
   \vspace{2mm}
   \centering BAD SLAM (298 frames)
   \vspace{1mm}
   \begin{minipage}{0.75\linewidth}
      \centering
      \includegraphics[width=0.8\linewidth]{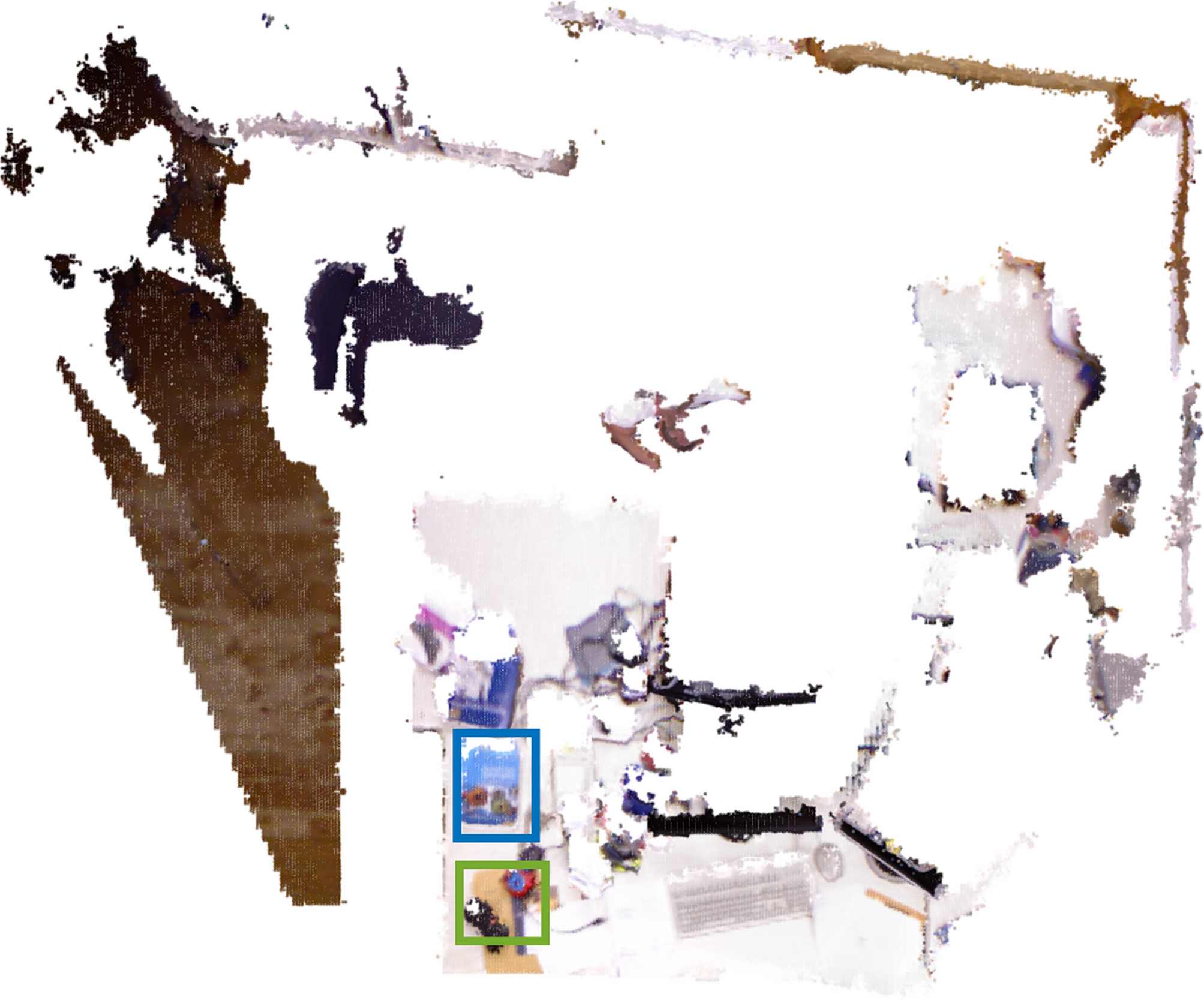}
   \end{minipage}%
   \begin{minipage}{0.2\linewidth}
   \centering
   \includegraphics[width=0.9\linewidth,cframe=mlgreen 1.5pt 0pt]{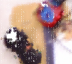}
   \includegraphics[width=0.9\linewidth,cframe=mlblue 1.5pt 0pt]{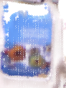}
   \end{minipage}
   \centering Ours (298 frames)
   \vspace*{1.5mm}
   \caption{Point cloud reconstruction on \emph{fr1/xyz}~\cite{sturm2012benchmark} by BAD SLAM~\cite{schops2019bad} (top) and our method (bottom).
    The proposed method reduces the blurry effect and recovers the clear texture. 
   }
   \label{fg:results_xyz}
\end{figure}

We use the first-order SH model to refine the poses together with surface geometry because the datasets satisfy the natural light assumption and compare against two baseline SLAM methods~\cite{bylow2013real, schops2019bad} and the method~\cite{Sommer2022gradient} which refines camera poses without PS loss. 
\begin{table}[t]
   \centering
   \begin{tabular}{l r r r r r r}
      & \rotatebox{90}{SDF-Fusion~\cite{bylow2013real}}
      & \rotatebox{90}{BAD SLAM~\cite{schops2019bad}}
      & \rotatebox{90}{gradient-SDF~\cite{Sommer2022gradient}}
      & \rotatebox{90}{\parbox{2.0cm}{Ours w/o pose refine}}
      & \rotatebox{90}{Ours (voxel)}
      & \rotatebox{90}{Ours} \\
      \hline
      fr1/xyz  & 2.3 & \textbf{1.8} & 2.0 & 2.1 & 2.0 & 2.0\\
      fr1/plant  & 4.3 & 1.9 & 11.2 & \textbf{1.5} & 2.5 & 2.5 \\
      fr3/household  & 4.0 & 1.5 & 1.5 & 2.8 & 1.1 & \textbf{0.7}\\
      fr2/desk  & - & 1.8 & 1.6 & 1.1 & 0.9 & \textbf{0.7}  \\
      fr2/rpy  & 2.2 & 0.9 & 7.0 & 1.3 & 0.6 & \textbf{0.4} \\
      fr2/xyz  & 1.8 & 1.3 & 1.3 & 1.0 & 0.9 & \textbf{0.6} \\
      \hline
   \end{tabular}
   \vspace{1.5mm}
   \caption{Root Mean Square Error (RMSE) of the absolute trajectory error (ATE) in cm on sequences from~\cite{sturm2012benchmark} compared against three baseline SLAM methods. From the right side, the third column represents the initial camera pose after tracking, the second column shows the error when modeling and optimizing the voxel center, and the last column shows the errors of the proposed method.}
   \label{tab:tracking_results}
\end{table}
We take maximum $300$ frames to initialize the SDF volume when the whole sequence length is more than $300$, then take $10$\% as keyframes using a sharpness detector~\cite{bansal2016} for the follow-up optimization. Table~\ref{tab:tracking_results} shows that the proposed method further improves the camera pose after the tracking phase and gets better accuracy than other state-of-the-art methods. We superior to~\cite{Sommer2022gradient} by introducing the image formation model to refine camera poses. We test directly optimizing~\eqref{eq:energy} on the voxel center to confirm that the step~\eqref{eq:surface_point} leads to more accurate results.  
The failing case (\emph{fr1/plant}) is due to absence of geometric information in the first few frames, causing tracking and PS optimization to fail. The results also indicate that PS method can deal with the wider baseline between two selected frames. Nevertheless, camera tracking, thus PS optimization might fail if the baseline is too large. 
The visualization of the refined surface compared to the initial surface also verified the improvement, see Figure~\ref{fg:results_xyz}. We obtain sharper and clearer textures compared to the state-of-the-art approaches. More visualization results are presented in the supplementary material. 

\subsection{Surface Geometry Improvement}
We evaluate the improvement of the surface geometry on both synthetic datasets and real-world datasets for two proposed image formation models: SH model and point light source (PLS) model. 
The synthetic data contains $90$ RGB images and corresponding depth images which are augmented with Kinect-like noise
~\cite{khoshelham2012accuracy}. We show the quantitative comparison on synthetic datasets and qualitative evaluation on real-world datasets. 
See supplementary material for more details.\par
Figure~\ref{fg:distance} shows the $3$ stage errors: error of initial point cloud, i.e., after camera tracking, error after voxel up-sampling and error after the optimization of the energy~\eqref{eq:energy}. To eliminate the influence of object size, as small object size leads to a small absolute point cloud distance error under the same voxel grid size, we compute the distance error and object size ratio as the measurement. The $x$-axis shows the point-to-point distance error $d_{c-c}$ w.r.t. the object size, i.e., $e = \frac{d_{c-c}}{d_{\text{max}}}$, where the $d_{\text{max}}$ is the point cloud bounding box size. The $y$-axis is the percentage of points: points number with error less than $e$ divided by the total points number of the point cloud. 
\begin{figure}[ht]
%
%
\definecolor{mycolor1}{rgb}{0.46600,0.67400,0.18800}%
\definecolor{mycolor2}{rgb}{0.00000,0.44700,0.74100}%
\definecolor{mycolor3}{rgb}{0.63500,0.07800,0.18400}%
\begin{tikzpicture}[scale=0.6]

\begin{axis}[%
width=3.192in,
height=2.128in,
at={(0.535in,1.298in)},
scale only axis,
unbounded coords=jump,
xmin=0,
xmax=0.03,
ymin=0,
ymax=100,
axis background/.style={fill=white},
axis x line*=bottom,
axis y line*=left,
xticklabel={$\pgfkeys{/pgf/number format/.cd,fixed,precision=3}\pgfmathprintnumber{\tick}\%$},
yticklabel={$\pgfmathprintnumber{\tick}\%$},
scaled x ticks = false,
xmajorgrids,
ymajorgrids,
legend style={at={(1.03,0.5)}, anchor=west, legend cell align=left, align=left, draw=white!15!black}
]
\addplot [color=white]
  table[row sep=crcr]{%
1	0.0  0.0\\
};
\addlegendentry{SH model}

\addplot [color=mycolor1, line width=1.0pt]
  table[row sep=crcr]{%
0	0.501346207408783\\
0.001001001001001	3.74152817751369\\
0.002002002002002	12.6543496425587\\
0.003003003003003	24.7145111874478\\
0.004004004004004	34.6393092563365\\
0.005005005005005	43.5428465323554\\
0.00600600600600601	51.6479435521307\\
0.00700700700700701	58.8431900473494\\
0.00800800800800801	65.1378702070374\\
0.00900900900900901	70.6341101104819\\
0.01001001001001	75.4526042150218\\
0.011011011011011	79.1755640144833\\
0.012012012012012	82.5828613870578\\
0.013013013013013	85.5166651193018\\
0.014014014014014	87.949122644137\\
0.015015015015015	89.5088664005199\\
0.016016016016016	91.0964627239811\\
0.017017017017017	92.4148175656856\\
0.018018018018018	93.4732151146597\\
0.019019019019019	94.448054962399\\
0.02002002002002	95.3022003527992\\
0.021021021021021	95.9520935846254\\
0.022022022022022	96.5741342493733\\
0.023023023023023	97.0290595116517\\
0.024024024024024	97.4932689629561\\
0.025025025025025	97.8553523349735\\
0.026026026026026	98.1431621947823\\
0.027027027027027	98.3752669204345\\
0.028028028028028	98.5609507009563\\
0.029029029029029	98.6816451582954\\
};
\addlegendentry{   initial pc}

\addplot [color=mycolor2, line width=1.0pt]
  table[row sep=crcr]{%
0	0.53961600539616\\
0.001001001001001	4.16275204162752\\
0.002002002002002	12.8929681289297\\
0.003003003003003	24.8632892486329\\
0.004004004004004	35.7616053576161\\
0.005005005005005	44.7038134470381\\
0.00600600600600601	52.6294235262942\\
0.00700700700700701	59.7889715978897\\
0.00800800800800801	65.898195658982\\
0.00900900900900901	71.5183927151839\\
0.01001001001001	76.2376237623762\\
0.011011011011011	80.4654188046542\\
0.012012012012012	83.9994218399942\\
0.013013013013013	86.7456818674568\\
0.014014014014014	89.0800028908\\
0.015015015015015	90.7157139071571\\
0.016016016016016	91.9491219194912\\
0.017017017017017	93.1126689311267\\
0.018018018018018	94.1726289417263\\
0.019019019019019	95.12900195129\\
0.02002002002002	95.9480619594806\\
0.021021021021021	96.5840379658404\\
0.022022022022022	97.0875189708752\\
0.023023023023023	97.5163209751632\\
0.024024024024024	97.8776709787767\\
0.025025025025025	98.2052949820529\\
0.026026026026026	98.4052419840524\\
0.027027027027027	98.5425549854255\\
0.028028028028028	98.6726409867264\\
0.029029029029029	98.7786369877864\\
};
\addlegendentry{   up-smapled pc}

\addplot [color=mycolor3, line width=1.0pt]
  table[row sep=crcr]{%
0	0.57420368518783\\
0.001001001001001	4.47936009141551\\
0.002002002002002	14.5493500928439\\
0.003003003003003	28.6330524210827\\
0.004004004004004	41.7997428938723\\
0.005005005005005	51.9011569775746\\
0.00600600600600601	60.897014712184\\
0.00700700700700701	68.01314097986\\
0.00800800800800801	73.8237394657906\\
0.00900900900900901	78.5202113983717\\
0.01001001001001	82.1225539208684\\
0.011011011011011	85.0564205113555\\
0.012012012012012	87.5046421939723\\
0.013013013013013	89.647193258106\\
0.014014014014014	91.346950435652\\
0.015015015015015	92.7696043422368\\
0.016016016016016	93.9208684473647\\
0.017017017017017	94.9150121411227\\
0.018018018018018	95.8263105270676\\
0.019019019019019	96.6490501356949\\
0.02002002002002	97.4546493358092\\
0.021021021021021	98.0917011855449\\
0.022022022022022	98.5544922153978\\
0.023023023023023	98.9144407941723\\
0.024024024024024	99.2315383516641\\
0.025025025025025	99.4115126410513\\
0.026026026026026	99.5629195829167\\
0.027027027027027	99.6429081559777\\
0.028028028028028	99.6857591772604\\
0.029029029029029	99.7400371375518\\
};
\addlegendentry{    refined pc}

\addplot [color=white]
  table[row sep=crcr]{%
1	0.0 0.0\\
};
\addlegendentry{PLS model}

\addplot [color=mycolor1, dashed, line width=1.0pt]
  table[row sep=crcr]{%
0	0.467683097932841\\
0.001001001001001	3.59180619212422\\
0.002002002002002	10.7567112524553\\
0.003003003003003	21.6443737723319\\
0.004004004004004	32.7752315031335\\
0.005005005005005	41.810868955196\\
0.00600600600600601	49.5931157047984\\
0.00700700700700701	56.935740342344\\
0.00800800800800801	63.2401085024787\\
0.00900900900900901	69.1048545505565\\
0.01001001001001	73.9594051070994\\
0.011011011011011	78.1872603124123\\
0.012012012012012	81.6948835469086\\
0.013013013013013	84.4074455149191\\
0.014014014014014	86.8393976241699\\
0.015015015015015	88.7475446637359\\
0.016016016016016	90.487325788046\\
0.017017017017017	92.068094659059\\
0.018018018018018	93.3869609952296\\
0.019019019019019	94.5187540922271\\
0.02002002002002	95.3699373304649\\
0.021021021021021	96.0621083154055\\
0.022022022022022	96.8197549340567\\
0.023023023023023	97.465157609204\\
0.024024024024024	97.9889626788888\\
0.025025025025025	98.3724628191937\\
0.026026026026026	98.6717800018707\\
0.027027027027027	98.9430361986718\\
0.028028028028028	99.1394630998036\\
0.029029029029029	99.3358900009354\\
};
\addlegendentry{   initial pc}

\addplot [color=mycolor2, dashed, line width=1.0pt]
  table[row sep=crcr]{%
0	0.401927371018921\\
0.001001001001001	3.34234340110471\\
0.002002002002002	11.2328123163709\\
0.003003003003003	22.9427664825479\\
0.004004004004004	34.0768597955106\\
0.005005005005005	43.6714067457986\\
0.00600600600600601	51.9591021271595\\
0.00700700700700701	59.1867434481138\\
0.00800800800800801	66.0007051357386\\
0.00900900900900901	71.7922200023504\\
0.01001001001001	76.8574450581737\\
0.011011011011011	81.0036432013163\\
0.012012012012012	84.489364202609\\
0.013013013013013	87.0983664355388\\
0.014014014014014	89.1479609824891\\
0.015015015015015	90.9484075684569\\
0.016016016016016	92.445645786814\\
0.017017017017017	93.7219414737337\\
0.018018018018018	94.727935127512\\
0.019019019019019	95.5882007286403\\
0.02002002002002	96.3544482312845\\
0.021021021021021	97.0689857797626\\
0.022022022022022	97.7482665413092\\
0.023023023023023	98.338230109296\\
0.024024024024024	98.7143025032319\\
0.025025025025025	99.0222117757668\\
0.026026026026026	99.2314020448936\\
0.027027027027027	99.4335409566341\\
0.028028028028028	99.5416617698907\\
0.029029029029029	99.6309789634505\\
};
\addlegendentry{   up-smapled pc}

\addplot [color=mycolor3, dashed, line width=1.0pt]
  table[row sep=crcr]{%
0	0.416759142807822\\
0.001001001001001	3.75576434613203\\
0.002002002002002	12.3794727626939\\
0.003003003003003	24.4630218736899\\
0.004004004004004	36.10515153757\\
0.005005005005005	45.6906118221499\\
0.00600600600600601	54.1614263520012\\
0.00700700700700701	61.9146260264852\\
0.00800800800800801	68.7726566545831\\
0.00900900900900901	74.2324480284087\\
0.01001001001001	79.2730142289956\\
0.011011011011011	83.2951098616557\\
0.012012012012012	86.7697467386748\\
0.013013013013013	89.5045744864492\\
0.014014014014014	91.2825824270672\\
0.015015015015015	92.661093437893\\
0.016016016016016	93.8176617099455\\
0.017017017017017	94.7646173953787\\
0.018018018018018	95.4575719464378\\
0.019019019019019	96.0888757367266\\
0.02002002002002	96.670858918399\\
0.021021021021021	97.2775024043797\\
0.022022022022022	97.928534438115\\
0.023023023023023	98.4636630416019\\
0.024024024024024	98.9913935537965\\
0.025025025025025	99.3859584227269\\
0.026026026026026	99.6350274962393\\
0.027027027027027	99.8323099307045\\
0.028028028028028	99.8988927523366\\
0.029029029029029	99.9235530566447\\
};
\addlegendentry{   refined pc}

\end{axis}
\end{tikzpicture}%
   \caption{The figure shows the distance error of the reconstruction at different stages of the proposed method.
    The percentage with error less than $0.010$\% of the point cloud size increases from $75.45$\% to $82.12$\% after optimization for SH model and from $73.95$\% to $79.27$\% for the PLS model.
    } \label{fg:distance}
\end{figure}
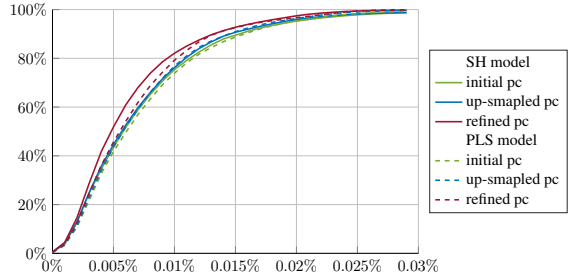

\begin{figure}[t]
   \centering
%
%
\definecolor{mycolor1}{rgb}{0.63500,0.07800,0.18400}%
\definecolor{mycolor2}{rgb}{0.46600,0.67400,0.18800}%
\definecolor{mycolor3}{rgb}{0.49400,0.18400,0.55600}%
\definecolor{mycolor4}{rgb}{0.92900,0.69400,0.12500}%
\definecolor{mycolor5}{rgb}{0.00000,0.44700,0.74100}%
\begin{tikzpicture}[scale=0.55]

\begin{axis}[%
width=3.495in,
height=2.33in,
at={(0.586in,1.197in)},
scale only axis,
xmin=0,
xmax=0.03,
ymin=0,
ymax=100,
axis background/.style={fill=white},
axis x line*=bottom,
axis y line*=left,
xticklabel={$\pgfkeys{/pgf/number format/.cd,fixed,precision=3}\pgfmathprintnumber{\tick}\%$},
yticklabel={$\pgfmathprintnumber{\tick}\%$},
scaled x ticks = false,
xmajorgrids,
ymajorgrids,
legend style={at={(1.03,0.5)}, anchor=west, legend cell align=left, align=left, draw=white!15!black}
]
\addplot [color=mycolor1, line width=1.0pt]
  table[row sep=crcr]{%
0	0.499043887878364\\
0.001001001001001	4.00167902616482\\
0.002002002002002	12.2195793106665\\
0.003003003003003	24.3878550440744\\
0.004004004004004	36.6913856629821\\
0.005005005005005	47.2039550394105\\
0.00600600600600601	56.19140898279\\
0.00700700700700701	63.8869455715685\\
0.00800800800800801	70.4211557296768\\
0.00900900900900901	75.8779907653561\\
0.01001001001001	80.518632526468\\
0.011011011011011	84.4270323212537\\
0.012012012012012	87.300032647731\\
0.013013013013013	89.7159647404505\\
0.014014014014014	91.6934844456882\\
0.015015015015015	93.330534956392\\
0.016016016016016	94.5618208105965\\
0.017017017017017	95.6765076255772\\
0.018018018018018	96.4927009001446\\
0.019019019019019	97.0850240194021\\
0.02002002002002	97.6027237535563\\
0.021021021021021	98.0224802947624\\
0.022022022022022	98.3676134508652\\
0.023023023023023	98.6707709528473\\
0.024024024024024	98.8713213003125\\
0.025025025025025	99.0625437246397\\
0.026026026026026	99.2537661489669\\
0.027027027027027	99.3936849960356\\
0.028028028028028	99.4776363042769\\
0.029029029029029	99.5382678046733\\
};
\addlegendentry{ours (SH)}

\addplot [color=mycolor2, dashed, line width=1.0pt]
  table[row sep=crcr]{%
0	0.384962038465651\\
0.001001001001001	3.44479919341287\\
0.002002002002002	11.1577886069568\\
0.003003003003003	23.0366172224683\\
0.004004004004004	34.452574815539\\
0.005005005005005	43.4380776340111\\
0.00600600600600601	51.2152273872993\\
0.00700700700700701	58.4363208628038\\
0.00800800800800801	65.0998304333878\\
0.00900900900900901	70.28459693558\\
0.01001001001001	74.5863949527199\\
0.011011011011011	78.7232092390889\\
0.012012012012012	82.4995035211806\\
0.013013013013013	85.7105757626678\\
0.014014014014014	88.1410305372665\\
0.015015015015015	90.04445394968\\
0.016016016016016	91.7416476986297\\
0.017017017017017	93.0783214433021\\
0.018018018018018	94.0682238279281\\
0.019019019019019	95.0275736698187\\
0.02002002002002	95.9853958845725\\
0.021021021021021	96.8851682681291\\
0.022022022022022	97.6932830234796\\
0.023023023023023	98.3104443867341\\
0.024024024024024	98.7381799850292\\
0.025025025025025	98.9367715128091\\
0.026026026026026	99.0330120224256\\
0.027027027027027	99.1002276164434\\
0.028028028028028	99.1414735491361\\
0.029029029029029	99.1827194818289\\
};
\addlegendentry{intrinsic3d~\cite{maier2017intrinsic3d}}

\addplot [color=mycolor3, dashed, line width=1.0pt]
  table[row sep=crcr]{%
0	0.376678676711431\\
0.001001001001001	3.02980674746151\\
0.002002002002002	9.52341958729119\\
0.003003003003003	19.1942351785129\\
0.004004004004004	27.7186374058303\\
0.005005005005005	35.5060596134949\\
0.00600600600600601	42.7694071405175\\
0.00700700700700701	49.3776613167376\\
0.00800800800800801	55.1834261382247\\
0.00900900900900901	61.0383229610219\\
0.01001001001001	66.0989190959712\\
0.011011011011011	70.6354405502784\\
0.012012012012012	74.7952833278742\\
0.013013013013013	78.5456927612185\\
0.014014014014014	81.6246315099902\\
0.015015015015015	84.3268915820505\\
0.016016016016016	87.0782836554209\\
0.017017017017017	89.207337045529\\
0.018018018018018	90.9023910907304\\
0.019019019019019	92.1552571241402\\
0.02002002002002	93.2771044873895\\
0.021021021021021	94.3580085162136\\
0.022022022022022	95.0867998689813\\
0.023023023023023	95.6681952178185\\
0.024024024024024	96.2250245660007\\
0.025025025025025	96.7982312479528\\
0.026026026026026	97.2240419259744\\
0.027027027027027	97.5433999344907\\
0.028028028028028	97.8545692761219\\
0.029029029029029	98.1247952833279\\
};
\addlegendentry{~\cite{bylow2019combining}}

\addplot [color=mycolor4, dashed, line width=1.0pt]
  table[row sep=crcr]{%
0	0.324388086110292\\
0.001001001001001	2.3296962547921\\
0.002002002002002	8.40460041285756\\
0.003003003003003	18.4016514302566\\
0.004004004004004	29.0769684458862\\
0.005005005005005	38.5137127690947\\
0.00600600600600601	46.5644352698319\\
0.00700700700700701	55.0280153347095\\
0.00800800800800801	62.9312887053966\\
0.00900900900900901	69.6254792096727\\
0.01001001001001	74.6387496313772\\
0.011011011011011	79.6225302270717\\
0.012012012012012	84.2524329106458\\
0.013013013013013	86.8475375995282\\
0.014014014014014	89.5311117664406\\
0.015015015015015	92.0967266293129\\
0.016016016016016	94.0430551459746\\
0.017017017017017	95.1636685343557\\
0.018018018018018	96.2842819227367\\
0.019019019019019	97.2279563550575\\
0.02002002002002	98.0241816573282\\
0.021021021021021	98.5255086994987\\
0.022022022022022	98.9678560896491\\
0.023023023023023	99.1742848717193\\
0.024024024024024	99.3807136537894\\
0.025025025025025	99.4986729578295\\
0.026026026026026	99.7051017398997\\
0.027027027027027	99.8820406959599\\
0.028028028028028	99.9115305219699\\
0.029029029029029	99.9115305219699\\
};
\addlegendentry{NeuS~\cite{wang2021neus}}

\addplot [color=mycolor5, dashed, line width=1.0pt]
  table[row sep=crcr]{%
0	0.532623169107856\\
0.001001001001001	3.40878828229028\\
0.002002002002002	11.4513981358189\\
0.003003003003003	23.9680426098535\\
0.004004004004004	35.6058588548602\\
0.005005005005005	45.0332889480692\\
0.00600600600600601	53.6884154460719\\
0.00700700700700701	62.1304926764314\\
0.00800800800800801	69.1877496671105\\
0.00900900900900901	73.7949400798935\\
0.01001001001001	77.4966711051931\\
0.011011011011011	80.3195739014647\\
0.012012012012012	83.3555259653795\\
0.013013013013013	85.6724367509987\\
0.014014014014014	87.5366178428762\\
0.015015015015015	89.5073235685752\\
0.016016016016016	90.7589880159787\\
0.017017017017017	91.664447403462\\
0.018018018018018	92.8628495339547\\
0.019019019019019	93.288948069241\\
0.02002002002002	94.1145139813582\\
0.021021021021021	94.8335552596538\\
0.022022022022022	95.9254327563249\\
0.023023023023023	96.6178428761651\\
0.024024024024024	97.3102529960053\\
0.025025025025025	97.7363515312916\\
0.026026026026026	98.2956058588549\\
0.027027027027027	98.7217043941411\\
0.028028028028028	98.9347536617843\\
0.029029029029029	99.2276964047936\\
};
\addlegendentry{volSDF~\cite{Yariv2021}}

\end{axis}
\end{tikzpicture}%
   \caption{Quantitative evaluation on synthetic datasets. The point number percentage for the points with distance error less than $0.015\%$ of point cloud size is $84.32\%$ (\cite{bylow2019combining}), $89.50\%$ (volSDF~\cite{Yariv2021}), $90.04\%$ (Intrinsic3d~\cite{maier2017intrinsic3d}) $92.09\%$ (NeuS~\cite{wang2021neus}), $93.33\%$ (ours)}\label{fg:compare} 
\end{figure}
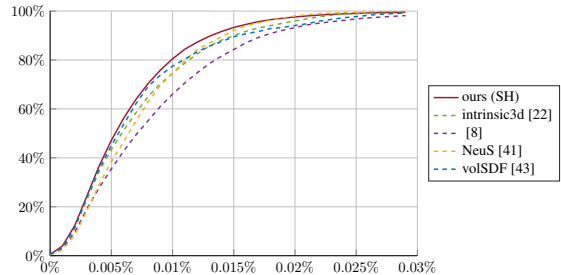

\begin{figure}[ht]
      \centering\includegraphics[width=0.22\linewidth, height=3.8cm]{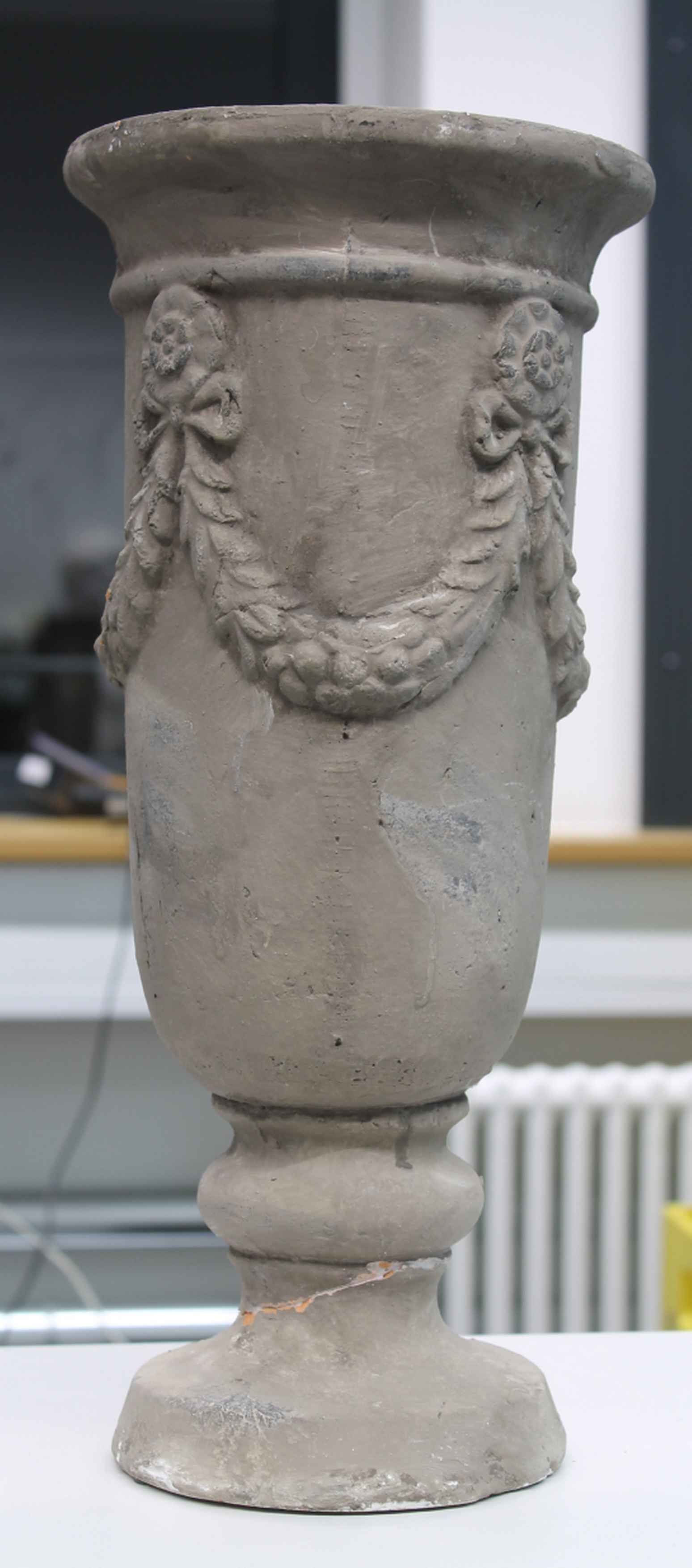}
      \hfill
      \centering \includegraphics[width=0.22\linewidth]{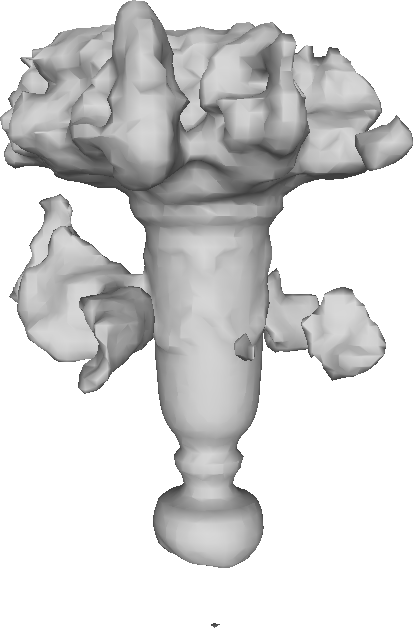}
      \hfill
      \centering \includegraphics[width=0.22\linewidth]{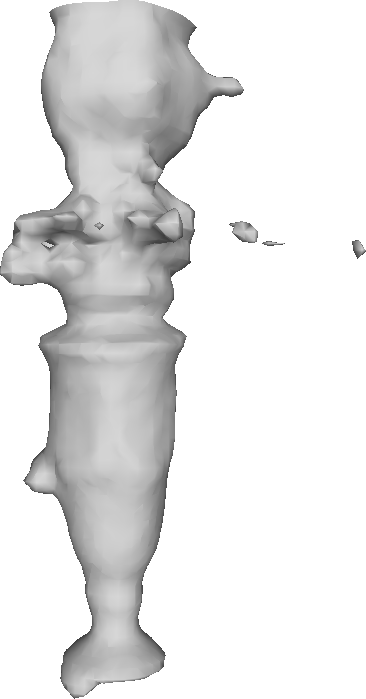}
      \hfill
      \centering \includegraphics[width=0.22\linewidth, height=3.8cm]{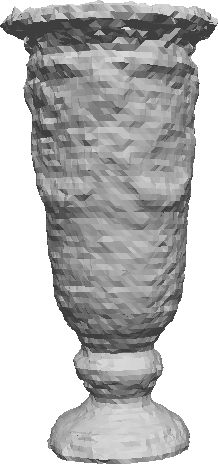} \\
      \makebox[0.22\linewidth][c]{\small{RGB image}}
      \hfill
      \makebox[0.22\linewidth][c]{\small{NeuS~\cite{wang2021neus}}}
      \hfill
      \makebox[0.22\linewidth][c]{\small{volSDF~\cite{Yariv2021}}}
      \hfill
      \makebox[0.22\linewidth][c]{\small{\cite{bylow2019combining}}} \\
      \vspace*{1.5mm}
      \centering 
      \includegraphics[width=0.2\linewidth]{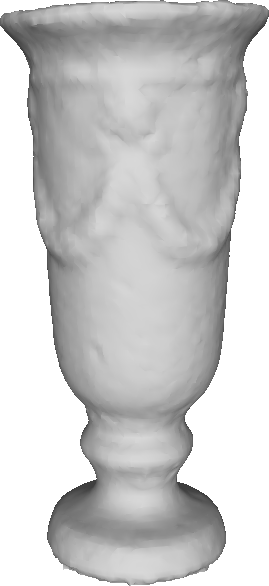} 
      \hfill
      \centering \includegraphics[width=0.2\linewidth, angle=180, origin=c,height=3.8cm]{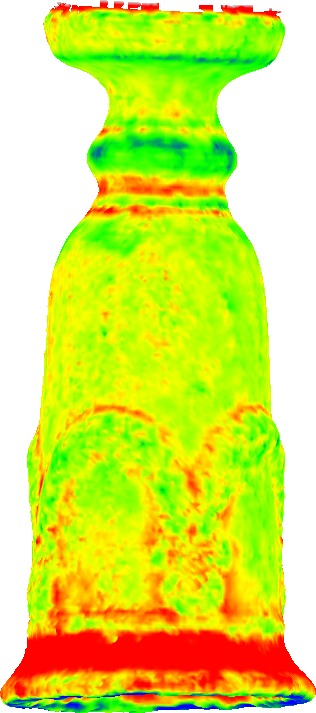}
      \hfill
      \centering \includegraphics[width=0.2\linewidth]{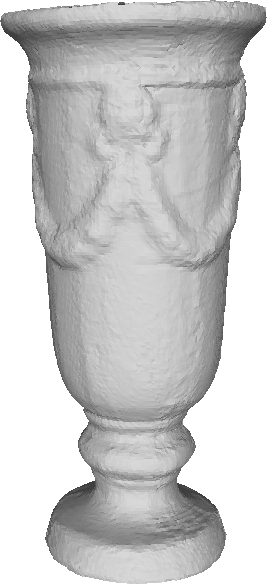}  
      \hfill
      \centering \includegraphics[width=0.2\linewidth, angle=180, origin=c,height=3.8cm]{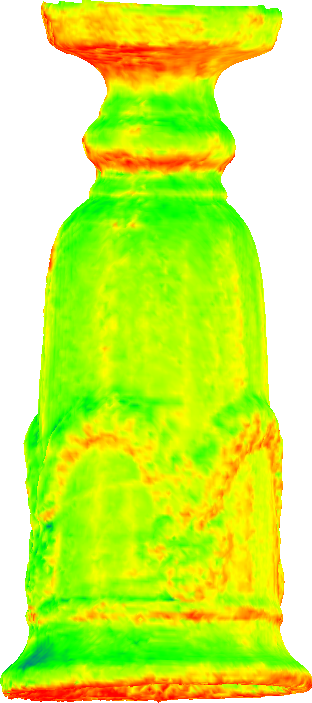} \\
      \vspace{0.15cm}
      \makebox[0.19\linewidth][c]{\small{Intrinsic3d~\cite{maier2017intrinsic3d}}}
      \hfill
      \makebox[0.19\linewidth][c]{\small{Intrinsic3d error}}
      \hfill
      \makebox[0.19\linewidth][c]{\small{Ours}}
      \hfill
      \makebox[0.19\linewidth][c]{\small{Ours error}}
      \vspace{1.5mm}
      \caption{Comparison with Intrinsic3d~\cite{maier2017intrinsic3d} on the vase dataset~\cite{zollhofer2015shading}. In the error map, from the yellow to red color transaction indicates a larger positive distance to the ground truth, and the blue direction indicates the negative distance. The standard deviation of our result and the laser scan is $4.5$mm, while~\cite{maier2017intrinsic3d} results in $5.8$mm.}\label{fg:compare_visual}
   \end{figure}

\paragraph*{SH model} Figure~\ref{fg:compare} shows the quantitative evaluation compared to the related methods~\cite{maier2017intrinsic3d, bylow2019combining, Yariv2021,wang2021neus} under same voxel size setting. The curve shows the proposed method performs best. 
Figure~\ref{fg:compare_visual} shows the visualization of the reconstructed error on the real-world datasets~\cite{zollhofer2015shading} where the laser-scanned ground truth is available. The neural rendering methods volSDF~\cite{Yariv2021}, NeuS~\cite{wang2021neus} perform well on the synthetic dataset but not well on the real-world datasets; even the mask setting is enabled in NeuS and the complex background setting is enabled in volSDF. 
For more experiments please refer to the supplementary material.

\begin{figure}[ht]
\begin{tabular}{c c}
   \includegraphics[width=0.45\linewidth, angle=180]{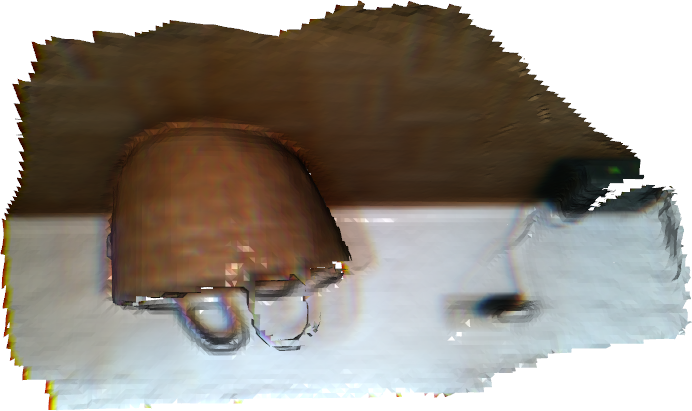} &
   \includegraphics[width=0.45\linewidth, angle=180]{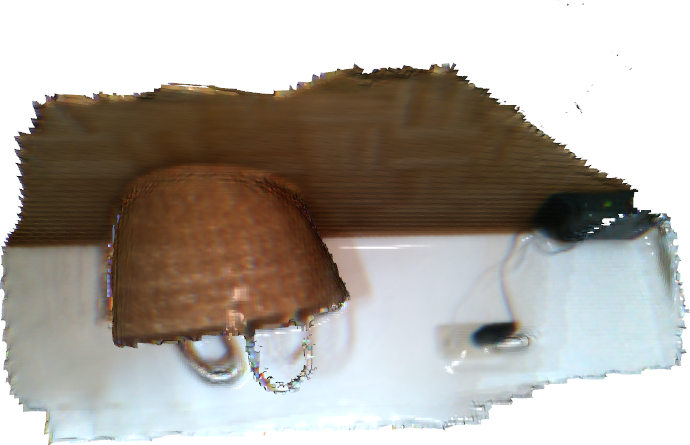} \\
   \includegraphics[width=0.42\linewidth, angle=180]{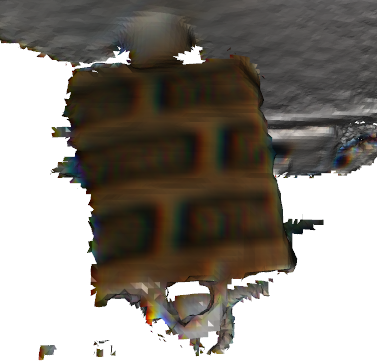} &
   \includegraphics[width=0.40\linewidth, angle=180]{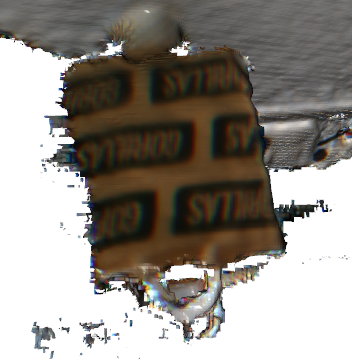} \\
   \centering initial reconstruction & \centering refined reconstruction
\end{tabular}
\vspace*{1.5mm}
\caption{The example reconstruction of a sequence concentrated on a single object. The sequence contains around $200$ RGB images and depth images that are recorded using the setting in Figure~\ref{fg:set_up}. 
}\label{fg:LED_results}
\end{figure}

\paragraph*{Point light source model} We do not find existing work performs a non-calibrated point light source method that enables a complete $3$D model. The work of Logothetis~\cite{logothetis2019differential} requires a specific setting and calibration, but the data and code are not publicly available. 
So we only present the visualization results of the datasets recorded and refined using the proposed setup mentioned in Section~\ref{se:modeling}. The sequences contain 
RGB and depth images of size $648\times 480$, $15\text{fps}$ frame rates. We plug in the sequences directly in our method without pre-processing. The quantitative and qualitative results are showed in Figure~\ref{fg:distance} and Figure~\ref{fg:LED_results}.

\section{Ablation Study}
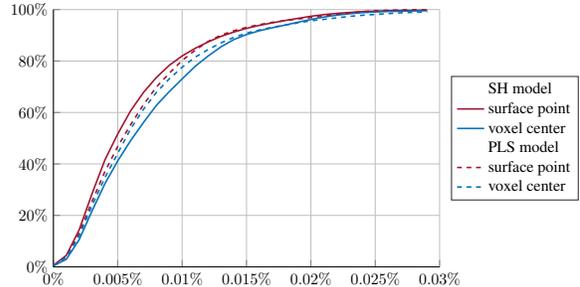
\begin{figure}[t]
   \centering
%
%
\definecolor{mycolor1}{rgb}{0.63500,0.07800,0.18400}%
\definecolor{mycolor2}{rgb}{0.00000,0.44700,0.74100}%
\begin{tikzpicture}[scale=0.6]

\begin{axis}[%
width=3.369in,
height=2.246in,
at={(0.565in,1.239in)},
scale only axis,
unbounded coords=jump,
xmin=0,
xmax=0.03,
ymin=0,
ymax=100,
axis background/.style={fill=white},
axis x line*=bottom,
axis y line*=left,
xticklabel={$\pgfkeys{/pgf/number format/.cd,fixed,precision=3}\pgfmathprintnumber{\tick}\%$},
yticklabel={$\pgfmathprintnumber{\tick}\%$},
scaled x ticks = false,
xmajorgrids,
ymajorgrids,
legend style={at={(1.03,0.5)}, anchor=west, legend cell align=left, align=left, draw=white!15!black}
]
\addplot [color=white]
  table[row sep=crcr]{%
1	0.0 0.0\\
};
\addlegendentry{SH model}

\addplot [color=mycolor1, line width=1.0pt]
  table[row sep=crcr]{%
0	0.57420368518783\\
0.001001001001001	4.43936580488502\\
0.002002002002002	14.4122268247393\\
0.003003003003003	28.461648335952\\
0.004004004004004	41.6140551349807\\
0.005005005005005	51.7240394229396\\
0.00600600600600601	60.7284673618055\\
0.00700700700700701	67.8588773032424\\
0.00800800800800801	73.6666190544208\\
0.00900900900900901	78.3745179260106\\
0.01001001001001	82.0111412655335\\
0.011011011011011	84.9392943865162\\
0.012012012012012	87.4217968861591\\
0.013013013013013	89.5757748893015\\
0.014014014014014	91.2669618625911\\
0.015015015015015	92.7210398514498\\
0.016016016016016	93.8837308955864\\
0.017017017017017	94.8521639765748\\
0.018018018018018	95.7748893015284\\
0.019019019019019	96.5890587058992\\
0.02002002002002	97.4089415797743\\
0.021021021021021	98.0488501642622\\
0.022022022022022	98.5316383373804\\
0.023023023023023	98.8801599771461\\
0.024024024024024	99.2201114126553\\
0.025025025025025	99.3972289672904\\
0.026026026026026	99.5543493786602\\
0.027027027027027	99.6371946864734\\
0.028028028028028	99.6857591772604\\
0.029029029029029	99.7371804027996\\
};
\addlegendentry{surface point}

\addplot [color=mycolor2, line width=1.0pt]
  table[row sep=crcr]{%
0	0.404760950587104\\
0.001001001001001	2.92950747405122\\
0.002002002002002	10.5718751252354\\
0.003003003003003	21.7508916763515\\
0.004004004004004	32.358834609065\\
0.005005005005005	41.2615717548992\\
0.00600600600600601	49.1323688534445\\
0.00700700700700701	56.1816214483229\\
0.00800800800800801	62.8201017913678\\
0.00900900900900901	68.1641485993668\\
0.01001001001001	73.0974231555324\\
0.011011011011011	77.9385244259207\\
0.012012012012012	81.9280246864105\\
0.013013013013013	85.4967338596561\\
0.014014014014014	88.350098184587\\
0.015015015015015	90.3258125275518\\
0.016016016016016	91.7865587304132\\
0.017017017017017	92.9467398709574\\
0.018018018018018	93.9866949865748\\
0.019019019019019	95.1168196208873\\
0.02002002002002	96.1828237085721\\
0.021021021021021	97.108564100509\\
0.022022022022022	97.8399390854807\\
0.023023023023023	98.4029976355548\\
0.024024024024024	98.8057548190598\\
0.025025025025025	99.0822746763916\\
0.026026026026026	99.2626137137819\\
0.027027027027027	99.3768284374624\\
0.028028028028028	99.5050695307177\\
0.029029029029029	99.597242816495\\
};
\addlegendentry{voxel center}

\addplot [color=white]
  table[row sep=crcr]{%
1	0.0 0.0\\
};
\addlegendentry{PLS model}

\addplot [color=mycolor1, dashed, line width=1.0pt]
  table[row sep=crcr]{%
0	0.458681660131686\\
0.001001001001001	4.00236738921358\\
0.002002002002002	13.0946215876304\\
0.003003003003003	25.4765603807551\\
0.004004004004004	37.0595053142956\\
0.005005005005005	46.8570442159256\\
0.00600600600600601	55.2958003501763\\
0.00700700700700701	63.120514907154\\
0.00800800800800801	69.9489531700821\\
0.00900900900900901	75.297773174521\\
0.01001001001001	80.2865527360608\\
0.011011011011011	84.1952109689034\\
0.012012012012012	87.6007989938596\\
0.013013013013013	90.0717614855367\\
0.014014014014014	91.793050726246\\
0.015015015015015	93.1074449458706\\
0.016016016016016	94.1727700919829\\
0.017017017017017	95.0482108949224\\
0.018018018018018	95.7263692633967\\
0.019019019019019	96.3478089319622\\
0.02002002002002	96.9199279919114\\
0.021021021021021	97.5635619343543\\
0.022022022022022	98.2022638159355\\
0.023023023023023	98.7151981455451\\
0.024024024024024	99.2059382012774\\
0.025025025025025	99.5314542181451\\
0.026026026026026	99.7533969569185\\
0.027027027027027	99.8742324480284\\
0.028028028028028	99.9161549653523\\
0.029029029029029	99.933417178368\\
};
\addlegendentry{surface point}

\addplot [color=mycolor2, dashed, line width=1.0pt]
  table[row sep=crcr]{%
0	0.515267175572519\\
0.001001001001001	3.72137404580153\\
0.002002002002002	12.0038167938931\\
0.003003003003003	23.7881679389313\\
0.004004004004004	34.7232824427481\\
0.005005005005005	44.236641221374\\
0.00600600600600601	53.6927480916031\\
0.00700700700700701	61.3263358778626\\
0.00800800800800801	67.8339694656489\\
0.00900900900900901	73.1202290076336\\
0.01001001001001	77.6240458015267\\
0.011011011011011	81.4980916030534\\
0.012012012012012	84.4274809160305\\
0.013013013013013	87.0419847328244\\
0.014014014014014	89.1984732824427\\
0.015015015015015	90.8778625954198\\
0.016016016016016	92.0801526717557\\
0.017017017017017	93.1488549618321\\
0.018018018018018	93.9885496183206\\
0.019019019019019	94.8473282442748\\
0.02002002002002	95.6679389312977\\
0.021021021021021	96.2022900763359\\
0.022022022022022	96.7843511450382\\
0.023023023023023	97.3568702290076\\
0.024024024024024	97.7958015267176\\
0.025025025025025	98.1202290076336\\
0.026026026026026	98.4446564885496\\
0.027027027027027	98.7595419847328\\
0.028028028028028	98.9312977099237\\
0.029029029029029	99.0553435114504\\
};
\addlegendentry{voxel center}

\end{axis}
\end{tikzpicture}%
   \caption{The figure shows the number of points with certain distance error range optimized on the voxel center (blue) and on surface points (red). Only less than $90.31$\% and $90.87$\% points have an error less than $0.015$\% percentage of the object size for the SH model and PLS model, while they increase to nearly $92.72$\% and $93.10$\% points when optimizing on the true surface points.}\label{fg:voxel_center}
\end{figure}

\paragraph*{Voxel center vs. surface point} To verify that formulating the image formation model on the surface point is theoretically more accurate and leads to better performance, we test the voxel-center formulation on the synthetic data as well. 
Figure~\ref{fg:voxel_center} shows that the surface point model results in more minor distance errors than the on-voxel-center approach. 
For camera pose refinement, the proposed method achieves the pose RMSE $6.3$cm for the SH model and $8.1$cm for the PLS model, while the on-voxel-center approach gives $7.8$cm and $15.9$cm, respectively.

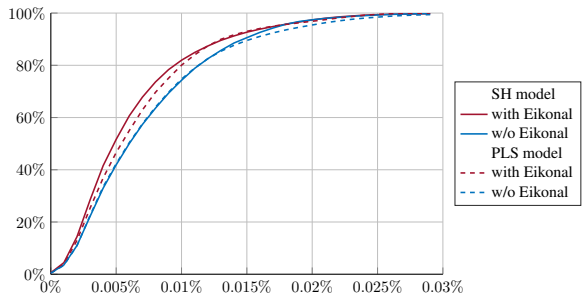
\begin{figure}[t]
   \centering
%
%
\definecolor{mycolor1}{rgb}{0.63500,0.07800,0.18400}%
\definecolor{mycolor2}{rgb}{0.00000,0.44700,0.74100}%
\begin{tikzpicture}[scale=0.6]

\begin{axis}[%
width=3.42in,
height=2.28in,
at={(0.574in,1.223in)},
scale only axis,
unbounded coords=jump,
xmin=0,
xmax=0.03,
ymin=0,
ymax=100,
axis background/.style={fill=white},
axis x line*=bottom,
axis y line*=left,
xticklabel={$\pgfkeys{/pgf/number format/.cd,fixed,precision=3}\pgfmathprintnumber{\tick}\%$},
yticklabel={$\pgfmathprintnumber{\tick}\%$},
scaled x ticks = false,
xmajorgrids,
ymajorgrids,
legend style={at={(1.03,0.5)}, anchor=west, legend cell align=left, align=left, draw=white!15!black}
]
\addplot [color=white]
  table[row sep=crcr]{%
1	0.0 0.0\\
};
\addlegendentry{SH model}

\addplot [color=mycolor1, line width=1.0pt]
  table[row sep=crcr]{%
0	0.568490215683474\\
0.001001001001001	4.43365233538066\\
0.002002002002002	14.3779460077132\\
0.003003003003003	28.430224253678\\
0.004004004004004	41.5683473789459\\
0.005005005005005	51.6840451364091\\
0.00600600600600601	60.6770461362662\\
0.00700700700700701	67.8017426081988\\
0.00800800800800801	73.6237680331381\\
0.00900900900900901	78.3430938437366\\
0.01001001001001	81.9768604485074\\
0.011011011011011	84.9221539780032\\
0.012012012012012	87.387516069133\\
0.013013013013013	89.5529210112841\\
0.014014014014014	91.2469647193258\\
0.015015015015015	92.7067561776889\\
0.016016016016016	93.8665904870733\\
0.017017017017017	94.8435937723182\\
0.018018018018018	95.7663190972718\\
0.019019019019019	96.5747750321383\\
0.02002002002002	97.3975146407656\\
0.021021021021021	98.0345664905013\\
0.022022022022022	98.525924867876\\
0.023023023023023	98.8773032423939\\
0.024024024024024	99.2201114126553\\
0.025025025025025	99.391515497786\\
0.026026026026026	99.5457791744037\\
0.027027027027027	99.631481216969\\
0.028028028028028	99.6829024425082\\
0.029029029029029	99.7343236680474\\
};
\addlegendentry{with Eikonal}

\addplot [color=mycolor2, line width=1.0pt]
  table[row sep=crcr]{%
0	0.469916152215977\\
0.001001001001001	3.57965539482171\\
0.002002002002002	10.9877453238736\\
0.003003003003003	22.1551644706533\\
0.004004004004004	32.9586289505206\\
0.005005005005005	41.9238920114254\\
0.00600600600600601	50.0460702110016\\
0.00700700700700701	57.3297705703492\\
0.00800800800800801	63.5907122454621\\
0.00900900900900901	69.3402745784576\\
0.01001001001001	74.3020363033263\\
0.011011011011011	78.8491661291809\\
0.012012012012012	82.5071408827052\\
0.013013013013013	85.6951994840136\\
0.014014014014014	88.52851746061\\
0.015015015015015	90.6062839767806\\
0.016016016016016	92.6379802819497\\
0.017017017017017	94.2965078780061\\
0.018018018018018	95.7661476089561\\
0.019019019019019	96.7842992720907\\
0.02002002002002	97.4246752050124\\
0.021021021021021	97.9544826315305\\
0.022022022022022	98.3599004883442\\
0.023023023023023	98.7514972818576\\
0.024024024024024	99.0094904634663\\
0.025025025025025	99.2720906661752\\
0.026026026026026	99.4886206578826\\
0.027027027027027	99.6084032064867\\
0.028028028028028	99.7373997972911\\
0.029029029029029	99.8111121348936\\
};
\addlegendentry{w/o Eikonal}

\addplot [color=white]
  table[row sep=crcr]{%
1	0.0 0.0\\
};
\addlegendentry{PLS model}

\addplot [color=mycolor1, dashed, line width=1.0pt]
  table[row sep=crcr]{%
0	0.453749599270055\\
0.001001001001001	3.95058075016646\\
0.002002002002002	12.9688540356588\\
0.003003003003003	25.3014722201672\\
0.004004004004004	36.8696209711228\\
0.005005005005005	46.6424995684447\\
0.00600600600600601	55.0319350940791\\
0.00700700700700701	62.9035042292422\\
0.00800800800800801	69.6653596705383\\
0.00900900900900901	75.0659663140243\\
0.01001001001001	80.0917363320263\\
0.011011011011011	84.0176567778846\\
0.012012012012012	87.4183127419792\\
0.013013013013013	89.9632561465808\\
0.014014014014014	91.6820793568593\\
0.015015015015015	93.0162018199305\\
0.016016016016016	94.1111193312125\\
0.017017017017017	94.9988902863061\\
0.018018018018018	95.6721165939188\\
0.019019019019019	96.283692140761\\
0.02002002002002	96.8656753224335\\
0.021021021021021	97.4895810214298\\
0.022022022022022	98.1455451160267\\
0.023023023023023	98.6609454760672\\
0.024024024024024	99.171413775246\\
0.025025025025025	99.5067939138369\\
0.026026026026026	99.7336687134719\\
0.027027027027027	99.866834356736\\
0.028028028028028	99.9112229044906\\
0.029029029029029	99.933417178368\\
};
\addlegendentry{with Eikonal}

\addplot [color=mycolor2, dashed, line width=1.0pt]
  table[row sep=crcr]{%
0	0.423728813559322\\
0.001001001001001	3.42362483102839\\
0.002002002002002	10.9779557034418\\
0.003003003003003	22.5044192575647\\
0.004004004004004	33.4979723406468\\
0.005005005005005	42.3624831028387\\
0.00600600600600601	50.4081314339191\\
0.00700700700700701	57.5647291255069\\
0.00800800800800801	63.9934491005511\\
0.00900900900900901	69.6916917957783\\
0.01001001001001	74.7738379952168\\
0.011011011011011	78.9669335551627\\
0.012012012012012	82.5335343662265\\
0.013013013013013	85.3774565872933\\
0.014014014014014	87.6572735780389\\
0.015015015015015	89.5341582614121\\
0.016016016016016	91.0912966621608\\
0.017017017017017	92.4352708744931\\
0.018018018018018	93.586877404596\\
0.019019019019019	94.5799105750234\\
0.02002002002002	95.4559633981491\\
0.021021021021021	96.3138192783612\\
0.022022022022022	97.0027035458043\\
0.023023023023023	97.6058022252262\\
0.024024024024024	98.0737236144328\\
0.025025025025025	98.4506602890714\\
0.026026026026026	98.8119995840699\\
0.027027027027027	99.079754601227\\
0.028028028028028	99.269522720183\\
0.029029029029029	99.409899136945\\
};
\addlegendentry{w/o Eikonal}

\end{axis}
\end{tikzpicture}%
   \caption{The number of points whose percentage distance error include the regularizer (red) and without the regularizer (blue). The points percentage of error less than $0.010$\% object size increase from $74.30$\% to $81.97$\% SH model and $74.77$\% to $80.09$\% for PLS model.}\label{fg:regularizer}
\end{figure}

\paragraph*{Eikonal constraint} We include Eikonal regularizer in~\eqref{eq:energy} in the multi-view PS energy. To verify the necessity of the regularizer not only from a theoretical but also from a heuristical point-of-view. We optimize our algorithm with and without Eikonal regularizer for SH and PLS models. As Figure~\ref{fg:regularizer} shows, Eikonal regularizer improves the accuracy of the results in both natural light and PLS settings.

\section{Conclusion and Future Work}
We perform camera pose tracking and $3$D surface recovery and refinement in a complete high-quality reconstruction pipeline using the gradient-SDF representation. We achieve good reconstruction quality and accurate pose tracking by enforcing the PS image formation model on the actual physical surface instead of the voxel center. 
Our method enables an easy and practical dense object $3$D reconstruction without pre-processing or any other calibrations in natural light or point light source scenarios. We demonstrate that our method achieves superior results quantitatively and qualitatively on both synthetic and real-world datasets. Yet, the limitation of the work is, it might fail for degenerated surface due to the absence of the geometry and shading information.
\par
In the future, we plan to include 
a general image formation model that can deal with non-Lambertian surfaces.

{\small
\bibliographystyle{ieee_fullname}
\bibliography{egbib}

\begin{thebibliography}{10}\itemsep=-1pt

\bibitem{bunny}
The stanford models.
\newblock \url{http:http://graphics.stanford.edu/data/3Dscanrep/}.
\newblock Accessed: 2021-09-19.

\bibitem{azinovic2021neural}
Dejan Azinovi{\'c}, Ricardo Martin-Brualla, Dan~B Goldman, Matthias
  Nie{\ss}ner, and Justus Thies.
\newblock Neural rgb-d surface reconstruction.
\newblock {\em arXiv preprint arXiv:2104.04532}, 2021.

\bibitem{bansal2016}
Raghav Bansal, Gaurav Raj, and Tanupriya Choudhury.
\newblock Blur image detection using laplacian operator and open-cv.
\newblock In {\em 2016 International Conference System Modeling and Advancement
  in Research Trends (SMART)}, pages 63--67, 2016.

\bibitem{basri2001}
R. Basri and D. Jacobs.
\newblock Photometric stereo with general, unknown lighting.
\newblock In {\em Proceedings of the 2001 IEEE Computer Society Conference on
  Computer Vision and Pattern Recognition. CVPR 2001}, volume~2, pages II--II,
  2001.

\bibitem{basri2003}
R. Basri and D.W. Jacobs.
\newblock Lambertian reflectance and linear subspaces.
\newblock {\em IEEE Transactions on Pattern Analysis and Machine Intelligence},
  25(2):218--233, 2003.

\bibitem{bjoern1996}
{\AA}ke Bj{\"o}rck.
\newblock {\em Numerical Methods for Least Squares Problems}.
\newblock Society for Industrial and Applied Mathematics, 1996.

\bibitem{brahimi2020springer}
M Brahimi, Y Quéau, B Haefner, and D Cremers.
\newblock {\em On the Well-Posedness of Uncalibrated Photometric Stereo Under
  General Lighting}, chapter Advances in Photometric 3D-Reconstruction, pages
  147--176.
\newblock Springer International Publishing, Cham, 2020.

\bibitem{bylow2019combining}
Erik Bylow, Robert Maier, Fredrik Kahl, and Carl Olsson.
\newblock Combining depth fusion and photometric stereo for fine-detailed 3d
  models.
\newblock In {\em Scandinavian Conference on Image Analysis}, pages 261--274.
  Springer, 2019.

\bibitem{bylow2013real}
Erik Bylow, J{\"u}rgen Sturm, Christian Kerl, Fredrik Kahl, and Daniel Cremers.
\newblock Real-time camera tracking and {3D} reconstruction using signed
  distance functions.
\newblock In {\em Robotics: Science and Systems}, volume~2, page~2, 2013.

\bibitem{gokhare2017review}
Vinod~G Gokhare, DN Raut, and DK Shinde.
\newblock A review paper on 3d-printing aspects and various processes used in
  the 3d-printing.
\newblock {\em Int. J. Eng. Res. Technol}, 6(06):953--958, 2017.

\bibitem{ha2021normalfusion}
Hyunho Ha, Joo~Ho Lee, Andreas Meuleman, and Min~H Kim.
\newblock Normalfusion: Real-time acquisition of surface normals for
  high-resolution rgb-d scanning.
\newblock In {\em Proceedings of the IEEE/CVF Conference on Computer Vision and
  Pattern Recognition}, pages 15970--15979, 2021.

\bibitem{haefner2019iccv}
B. Haefner, Z. Ye, M. Gao, T. Wu, Y. Quéau, and D. Cremers.
\newblock Variational uncalibrated photometric stereo under general lighting.
\newblock In {\em International Conference on Computer Vision (ICCV)}, Seoul,
  South Korea, October 2019.

\bibitem{he2021towards}
Lingzhi He, Hongguang Zhu, Feng Li, Huihui Bai, Runmin Cong, Chunjie Zhang,
  Chunyu Lin, Meiqin Liu, and Yao Zhao.
\newblock Towards fast and accurate real-world depth super-resolution:
  Benchmark dataset and baseline.
\newblock In {\em Proceedings of the IEEE/CVF Conference on Computer Vision and
  Pattern Recognition (CVPR)}, pages 9229--9238, June 2021.

\bibitem{hull2007}
Jonathan~J. Hull, Berna Erol, Jamey Graham, Qifa Ke, Hidenobu Kishi, Jorge
  Moraleda, and Daniel~G. Van~Olst.
\newblock Paper-based augmented reality.
\newblock In {\em 17th International Conference on Artificial Reality and
  Telexistence (ICAT 2007)}, pages 205--209, 2007.

\bibitem{Kajiya1986}
James~T. Kajiya.
\newblock The rendering equation.
\newblock In {\em Proceedings of the 13th Annual Conference on Computer
  Graphics and Interactive Techniques}, SIGGRAPH '86, page 143–150, New York,
  NY, USA, 1986. Association for Computing Machinery.

\bibitem{khoshelham2012accuracy}
Kourosh Khoshelham and Sander~Oude Elberink.
\newblock Accuracy and resolution of {Kinect} depth data for indoor mapping
  applications.
\newblock {\em Sensors}, 12(2):1437--1454, 2012.

\bibitem{lee2020texturefusion}
Joo~Ho Lee, Hyunho Ha, Yue Dong, Xin Tong, and Min~H Kim.
\newblock Texturefusion: High-quality texture acquisition for real-time rgb-d
  scanning.
\newblock In {\em Proceedings of the IEEE/CVF Conference on Computer Vision and
  Pattern Recognition}, pages 1272--1280, 2020.

\bibitem{lin2021barf}
Chen-Hsuan Lin, Wei-Chiu Ma, Antonio Torralba, and Simon Lucey.
\newblock Barf: Bundle-adjusting neural radiance fields.
\newblock In {\em IEEE International Conference on Computer Vision ({ICCV})},
  2021.

\bibitem{Logothetis2018}
Fotios Logothetis, Roberto Mecca, and Roberto Cipolla.
\newblock A differential volumetric approach to multi-view photometric stereo,
  2018.

\bibitem{logothetis2019differential}
Fotios Logothetis, Roberto Mecca, and Roberto Cipolla.
\newblock A differential volumetric approach to multi-view photometric stereo.
\newblock In {\em Proceedings of the IEEE/CVF International Conference on
  Computer Vision}, pages 1052--1061, 2019.

\bibitem{lorensen1987marching}
William~E Lorensen and Harvey~E Cline.
\newblock Marching cubes: A high resolution 3d surface construction algorithm.
\newblock {\em ACM siggraph computer graphics}, 21(4):163--169, 1987.

\bibitem{maier2017intrinsic3d}
R. Maier, K. Kim, D. Cremers, J. Kautz, and M. Niessner.
\newblock Intrinsic3d: High-quality {3D} reconstruction by joint appearance and
  geometry optimization with spatially-varying lighting.
\newblock In {\em International Conference on Computer Vision (ICCV)}, Venice,
  Italy, October 2017.

\bibitem{Mecca2014}
Roberto Mecca, Aaron Wetzler, Alfred~M. Bruckstein, and Ron Kimmel.
\newblock Near field photometric stereo with point light sources.
\newblock {\em SIAM Journal on Imaging Sciences}, 7.

\bibitem{mildenhall2020nerf}
Ben Mildenhall, Pratul~P. Srinivasan, Matthew Tancik, Jonathan~T. Barron, Ravi
  Ramamoorthi, and Ren Ng.
\newblock Nerf: Representing scenes as neural radiance fields for view
  synthesis.
\newblock In {\em ECCV}, 2020.

\bibitem{newcombe2011kinectfusion}
Richard~A Newcombe, Shahram Izadi, Otmar Hilliges, David Molyneaux, David Kim,
  Andrew~J Davison, Pushmeet Kohi, Jamie Shotton, Steve Hodges, and Andrew
  Fitzgibbon.
\newblock Kinect{F}usion: Real-time dense surface mapping and tracking.
\newblock In {\em 10th IEEE International Symposium on Mixed and Augmented
  Reality}, pages 127--136. IEEE, 2011.

\bibitem{niessner2013real}
Matthias Nie{\ss}ner, Michael Zollh{\"o}fer, Shahram Izadi, and Marc
  Stamminger.
\newblock Real-time {3D} reconstruction at scale using voxel hashing.
\newblock {\em ACM Transactions on Graphics (ToG)}, 32(6):1--11, 2013.

\bibitem{Osher2003}
Stanley~J. Osher and Ronald Fedkiw.
\newblock {\em Level set methods and dynamic implicit surfaces.}, volume 153 of
  {\em Applied mathematical sciences}.
\newblock Springer, 2003.

\bibitem{peng2017}
S. Peng, B. Haefner, Y. Quéau, and D. Cremers.
\newblock Depth super-resolution meets uncalibrated photometric stereo.
\newblock In {\em {International Conference on Computer Vision Workshops
  (ICCVW)}}, 2017.

\bibitem{peng2020convolution}
Songyou Peng, Michael Niemeyer, Lars Mescheder, Marc Pollefeys, and Andreas
  Geiger.
\newblock Convolutional occupancy networks, 2020.

\bibitem{Queau2017}
Yvain Quéau, Bastien Durix, Tao Wu, Daniel Cremers, François Lauze, and
  Jean-Denis Durou.
\newblock Led-based photometric stereo: Modeling, calibration and numerical
  solution, 2017.

\bibitem{jorge2018}
Jorge Reyna, Jose Hanham, and Peter Meier.
\newblock The internet explosion, digital media principles and implications to
  communicate effectively in the digital space.
\newblock {\em E-Learning and Digital Media}, 15(1):36--52, 2018.

\bibitem{sang2020wacv}
L. Sang, B. Haefner, and D. Cremers.
\newblock Inferring super-resolution depth from a moving light-source enhanced
  {RGB-D} sensor: A variational approach.
\newblock In {\em IEEE Winter Conference on Applications of Computer Vision
  (WACV)}, Colorado, USA, March 2020.

\bibitem{yu2021plenoxels}
{Sara Fridovich-Keil and Alex Yu}, Matthew Tancik, Qinhong Chen, Benjamin
  Recht, and Angjoo Kanazawa.
\newblock Plenoxels: Radiance fields without neural networks.
\newblock In {\em CVPR}, 2022.

\bibitem{schoenberger2016sfm}
Johannes~Lutz Sch\"{o}nberger and Jan-Michael Frahm.
\newblock {Structure-from-Motion Revisited}.
\newblock In {\em Conference on Computer Vision and Pattern Recognition
  (CVPR)}, 2016.

\bibitem{schoenberger2016mvs}
Johannes~Lutz Sch\"{o}nberger, Enliang Zheng, Marc Pollefeys, and Jan-Michael
  Frahm.
\newblock {Pixelwise View Selection for Unstructured Multi-View Stereo}.
\newblock In {\em European Conference on Computer Vision (ECCV)}, 2016.

\bibitem{schops2019bad}
Thomas Schops, Torsten Sattler, and Marc Pollefeys.
\newblock Bad slam: Bundle adjusted direct rgb-d slam.
\newblock In {\em Proceedings of the IEEE/CVF Conference on Computer Vision and
  Pattern Recognition}, pages 134--144, 2019.

\bibitem{Soltani2017}
Amir~Arsalan Soltani, Haibin Huang, Jiajun Wu, Tejas~D Kulkarni, and Joshua~B
  Tenenbaum.
\newblock Synthesizing 3d shapes via modeling multi-view depth maps and
  silhouettes with deep generative networks.
\newblock In {\em Proceedings of the IEEE Conference on Computer Vision and
  Pattern Recognition}, pages 1511--1519, 2017.

\bibitem{Sommer2022gradient}
C Sommer, L Sang, D Schubert, and D Cremers.
\newblock Gradient-{SDF}: {A} semi-implicit surface representation for 3d
  reconstruction.
\newblock In {\em IEEE Conference on Computer Vision and Pattern Recognition
  (CVPR)}, 2022.

\bibitem{sturm2012benchmark}
J{\"u}rgen Sturm, Nikolas Engelhard, Felix Endres, Wolfram Burgard, and Daniel
  Cremers.
\newblock A benchmark for the evaluation of rgb-d slam systems.
\newblock In {\em 2012 IEEE/RSJ international conference on intelligent robots
  and systems}, pages 573--580. IEEE, 2012.

\bibitem{turk1994zippered}
Greg Turk and Marc Levoy.
\newblock Zippered polygon meshes from range images.
\newblock In {\em Proceedings of the 21st Annual Conference on Computer
  Graphics and Interactive Techniques}, SIGGRAPH '94, page 311–318, New York,
  NY, USA, 1994. Association for Computing Machinery.

\bibitem{wang2021neus}
Peng Wang, Lingjie Liu, Yuan Liu, Christian Theobalt, Taku Komura, and Wenping
  Wang.
\newblock Neus: Learning neural implicit surfaces by volume rendering for
  multi-view reconstruction, 2021.

\bibitem{Woodham1977Reflectance}
Robert~J. Woodham.
\newblock Reflectance map techniques for analyzing surface defects in metal
  castings.
\newblock 1977.

\bibitem{Yariv2021}
Lior Yariv, Jiatao Gu, Yoni Kasten, and Yaron Lipman.
\newblock Volume rendering of neural implicit surfaces, 2021.

\bibitem{yu2021plenoctrees}
Alex Yu, Ruilong Li, Matthew Tancik, Hao Li, Ren Ng, and Angjoo Kanazawa.
\newblock {PlenOctrees} for real-time rendering of neural radiance fields.
\newblock In {\em ICCV}, 2021.

\bibitem{physg2020}
Kai Zhang, Fujun Luan, Qianqian Wang, Kavita Bala, and Noah Snavely.
\newblock Physg: Inverse rendering with spherical gaussians for physics-based
  material editing and relighting.
\newblock In {\em The IEEE/CVF Conference on Computer Vision and Pattern
  Recognition (CVPR)}, 2021.

\bibitem{zhou2014color}
Qian-Yi Zhou and Vladlen Koltun.
\newblock Color map optimization for 3d reconstruction with consumer depth
  cameras.
\newblock {\em ACM Trans. Graph.}, 33(4), jul 2014.

\bibitem{zollhofer2015shading}
Michael Zollh{\"o}fer, Angela Dai, Matthias Innmann, Chenglei Wu, Marc
  Stamminger, Christian Theobalt, and Matthias Nie{\ss}ner.
\newblock Shading-based refinement on volumetric signed distance functions.
\newblock {\em ACM Transactions on Graphics (TOG)}, 34(4):1--14, 2015.

\end{thebibliography}
}

\onecolumn
\appendix


\setcounter{equation}{0}
\renewcommand{\theequation}{A\arabic{equation}}
\setcounter{table}{0}
\renewcommand{\thetable}{A\arabic{table}}
\setcounter{figure}{0}
\renewcommand{\thefigure}{A\arabic{figure}}

\section*{Supplementary Material}
This Supplement contains information on code, datasets, some more visualizations of results, and the mathematical details of the proposed algorithm.  

\section{Used Code and Datasets}

The following table summarizes the code and datasets we use for evaluation and comparison. We compare with the methods whose code is publicly available and can be successfully compiled and run. Only for the work of Bylow el at.~\cite{bylow2019combining} we implement their method ourselves because their method has similarities with our proposed method, but the author has not published the code yet. Our code and recorded datasets will be made publicly available after the publication.

\begin{table}[H]
    \centering
    \begin{longtable}{l p{4cm} l l p{5cm} l}
    \toprule
         && code/data & year & link & license \\
    \midrule
        \cite{sturm2012benchmark} & TUM RGB-D Benchmark & dataset & 2012 &  {\small \url{https://vision.in.tum.de/data/datasets/rgbd-dataset}} &  CC BY 4.0 \\
        \cite{maier2017intrinsic3d} & Intrinsic3D Dataset & dataset & 2017 & \small{\url{https://vision.in.tum.de/data/datasets/intrinsic3d}} & CC BY 4.0 \\
        \cite{zollhofer2015shading} & multi-view stereo dataset & dataset & 2015 & {\small\url{http://graphics.stanford.edu/projects/vsfs/}} & CC BY-NC-SA 4.0 \\
        \cite{zhou2014color} & 3D Scene Data & dataset & 2014 & \small{\url{https://qianyi.info/scenedata.html}} & - \\
        \cite{sturm2012benchmark} & TUM RGB-D Benchmark & code & 2012 & {\small\url{https://vision.in.tum.de/data/datasets/rgbd-dataset/tools}} & BSD-2 \\
        \cite{schops2019bad} & BAD SLAM & code & 2019 & {\small\url{https://github.com/ETH3D/badslam}} & BSD-3 \\
        \cite{Sommer2022gradient} & Gradient-SDF & code & 2022 & {\small\url{https://github.com/c-sommer/gradient-sdf}} & BSD-3 \\
        \cite{maier2017intrinsic3d} & intrinsic3d & code & 2017 & {\small\url{https://github.com/NVlabs/intrinsic3d}} & BSD-3 \\
        \cite{wang2021neus} & NeuS & code & 2021 & {\small\url{https://github.com/Totoro97/NeuS}} & MIT License \\
        \cite{Yariv2021} & volSDF & code & 2021 & {\small\url{https://github.com/lioryariv/volsdf}} & MIT License \\
    \bottomrule
    \end{longtable}
    \vspace{0.25cm}
    \caption{Used datasets and code in our submission, together with reference, link, and license.}
    \label{tab:code_data}
\end{table}


\section{Visualization Results on TUM RGB-D Datasets}
\begin{figure}[h]
    \makebox[0.33\linewidth][c]{\small{Gradient-SDF~\cite{Sommer2022gradient}}}
    \hfill
    \centering
    \makebox[0.33\linewidth][c]{\small{BAD SLAM~\cite{schops2019bad}}}
    \hfill
    \centering
    \makebox[0.33\linewidth][c]{\small{Ours}}
    \\
    \centering
    \includegraphics[height=0.3\linewidth, angle=90, origin=c]{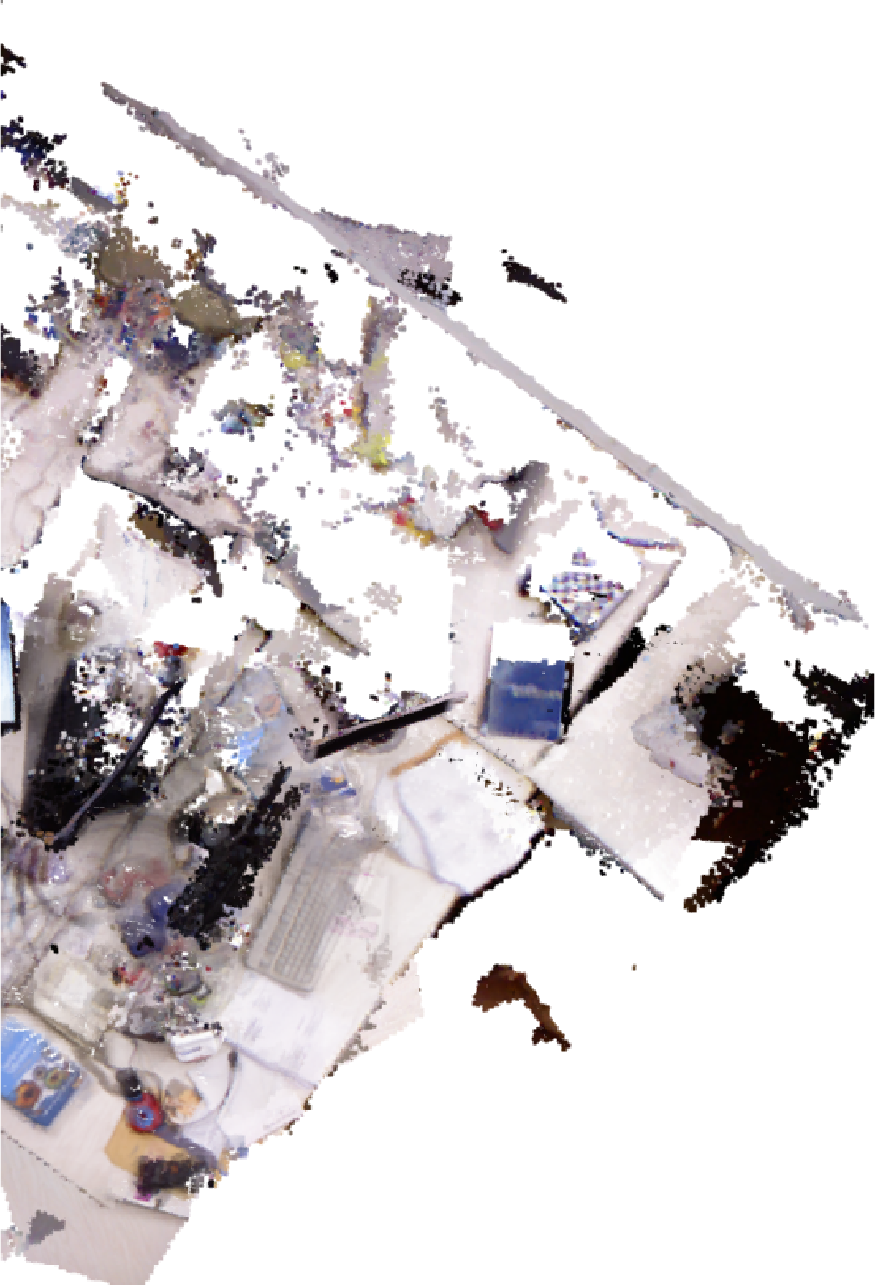} 
    \hfill 
    \includegraphics[height=0.3\linewidth, angle=90, origin=c]{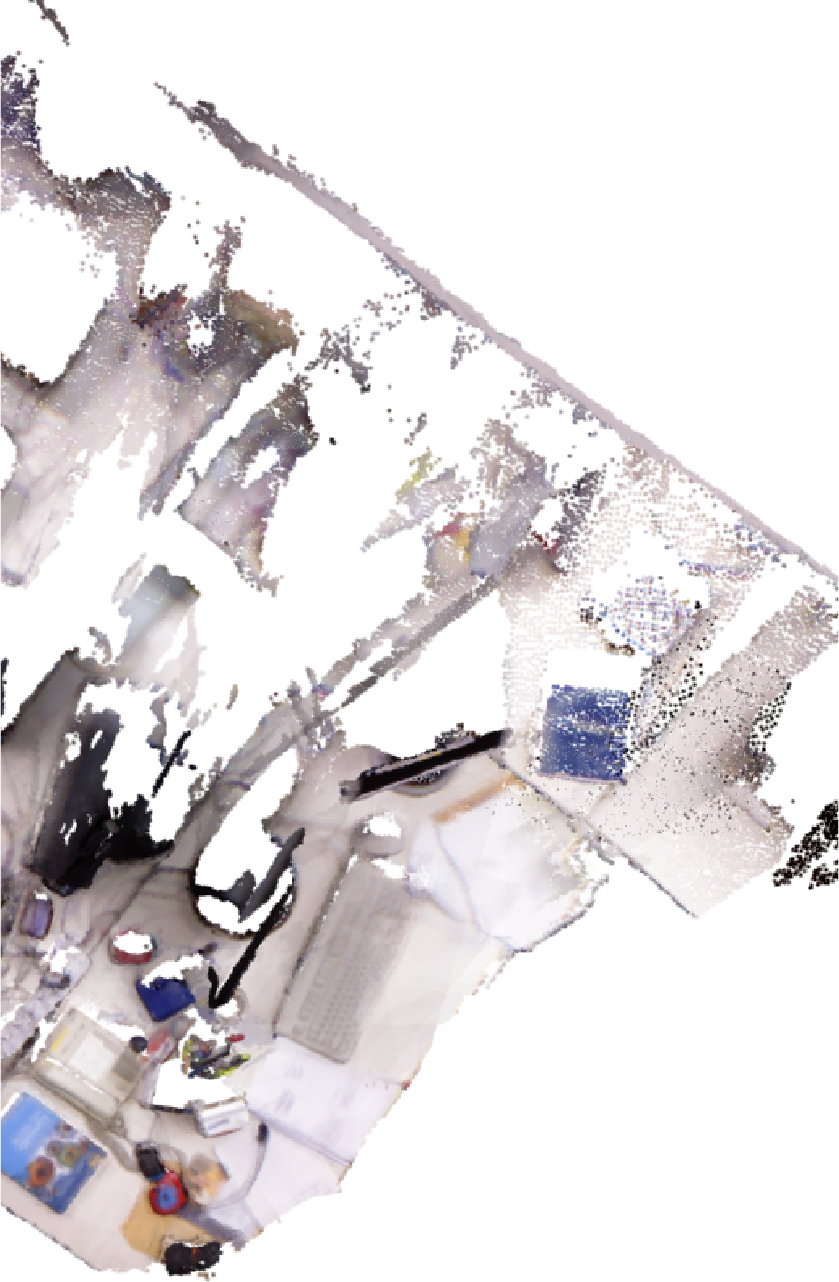}
    \hfill
    \includegraphics[height=0.3\linewidth, angle=90, origin=c]{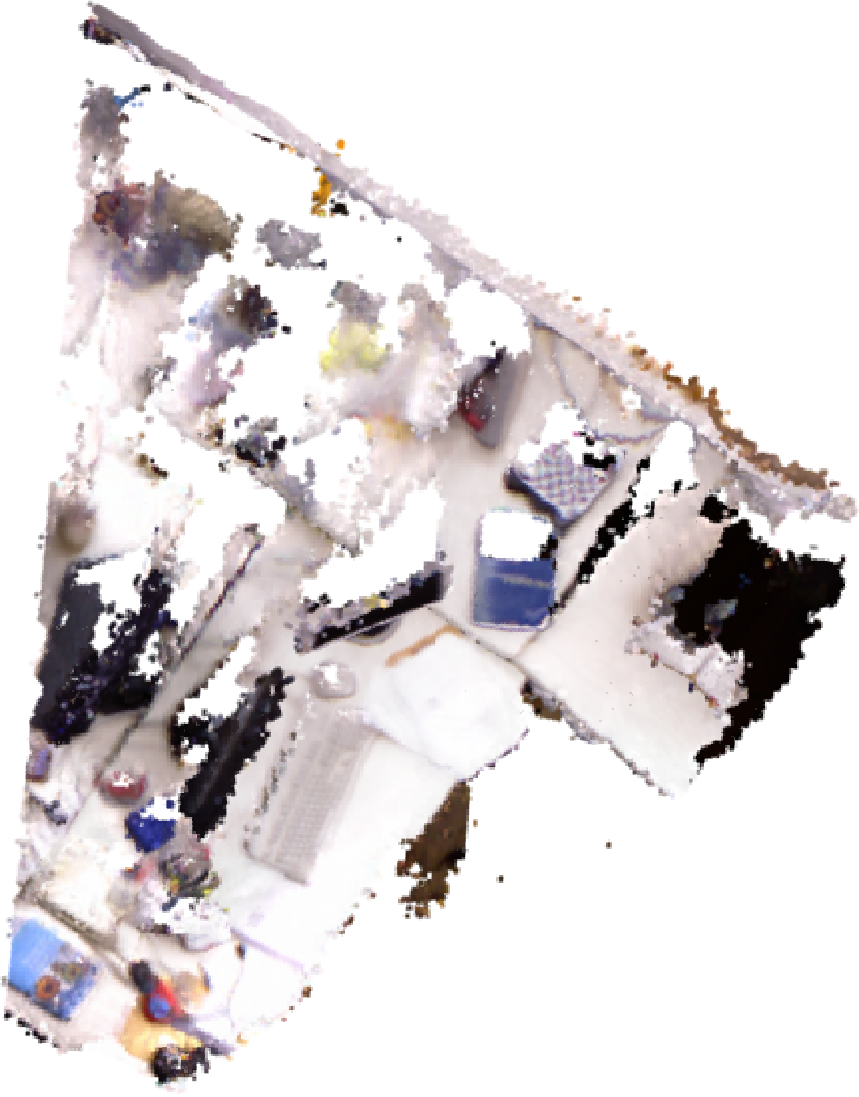}
    \\ \vspace*{0.25cm}
    \includegraphics[width=0.33\linewidth]{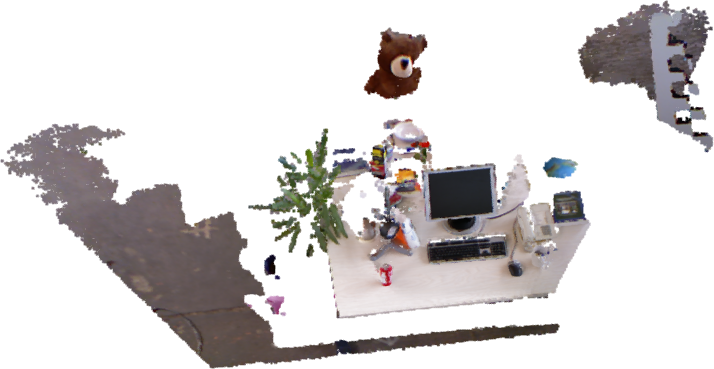} 
    \hfill
    \includegraphics[width=0.33\linewidth]{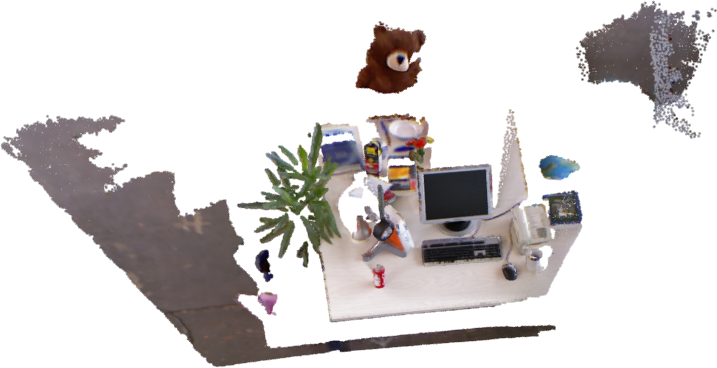} 
    \hfill
    \includegraphics[width=0.33\linewidth]{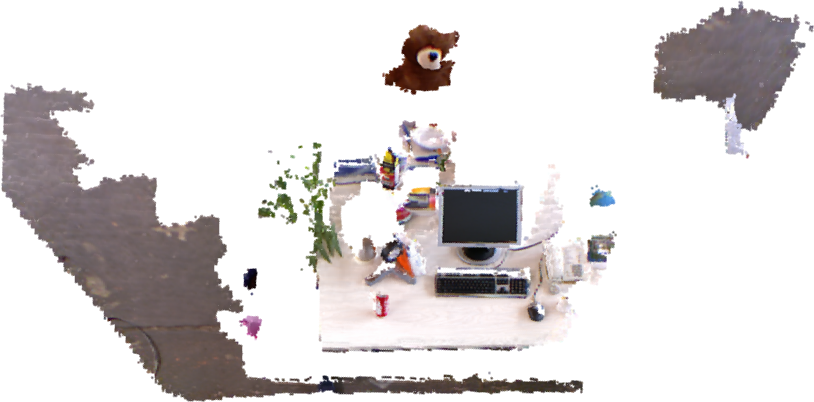} 
    \\ \vspace*{0.25cm}
    \includegraphics[width=0.33\linewidth,angle=180]{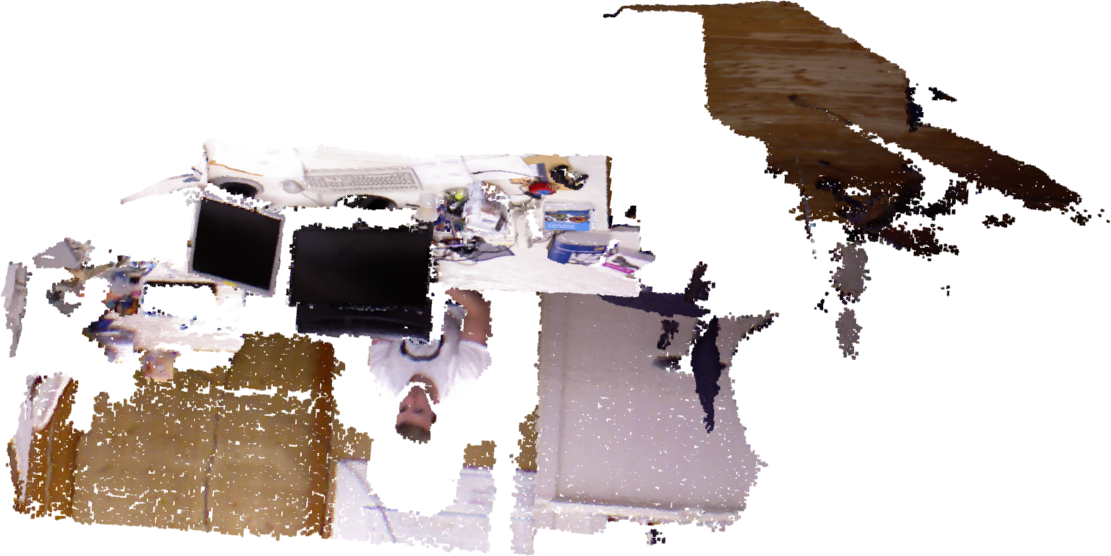}
    \hfill
    \includegraphics[width=0.33\linewidth, angle=180]{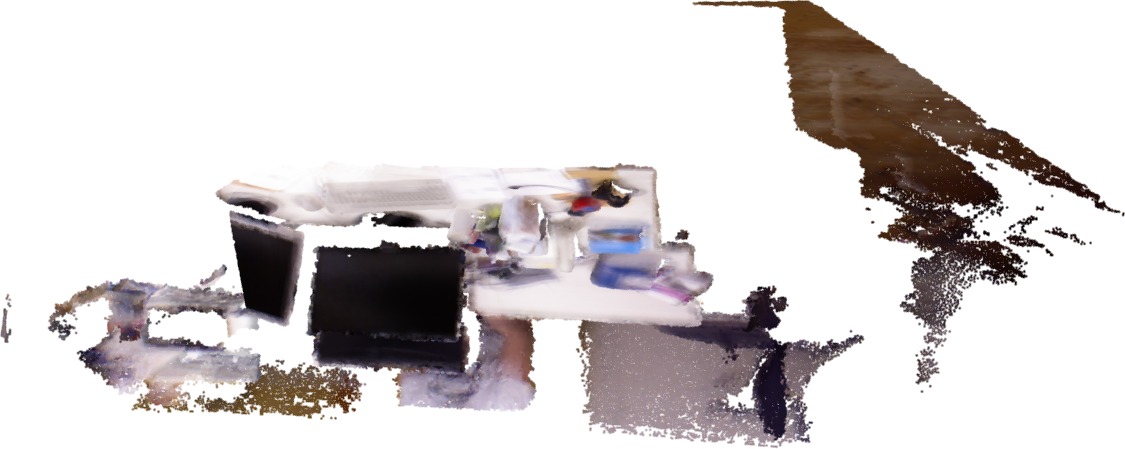}
    \hfill
    \includegraphics[width=0.33\linewidth, angle=180]{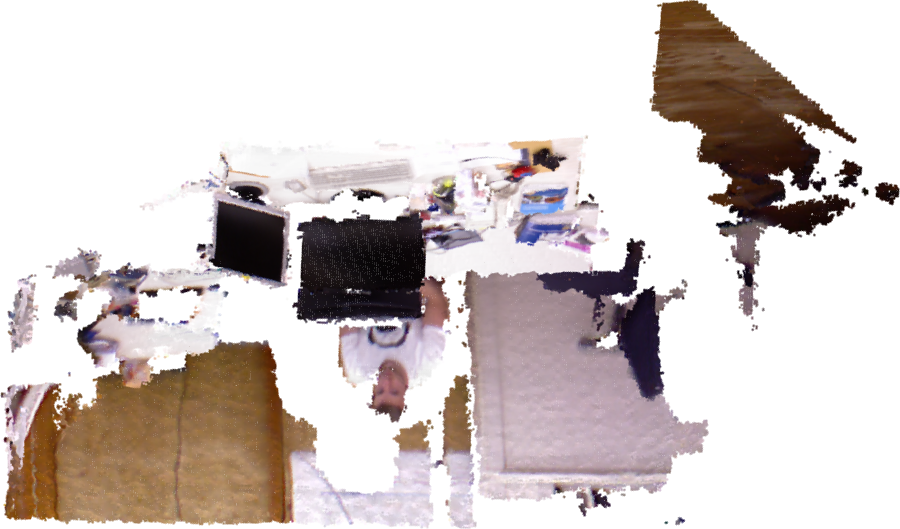} 
    \\ \vspace*{0.35cm}
    \includegraphics[width=0.33\linewidth,angle=180]{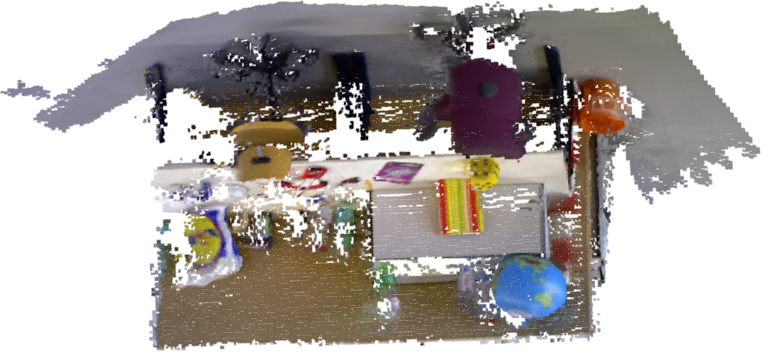}
    \hfill
    \includegraphics[width=0.33\linewidth, angle=180]{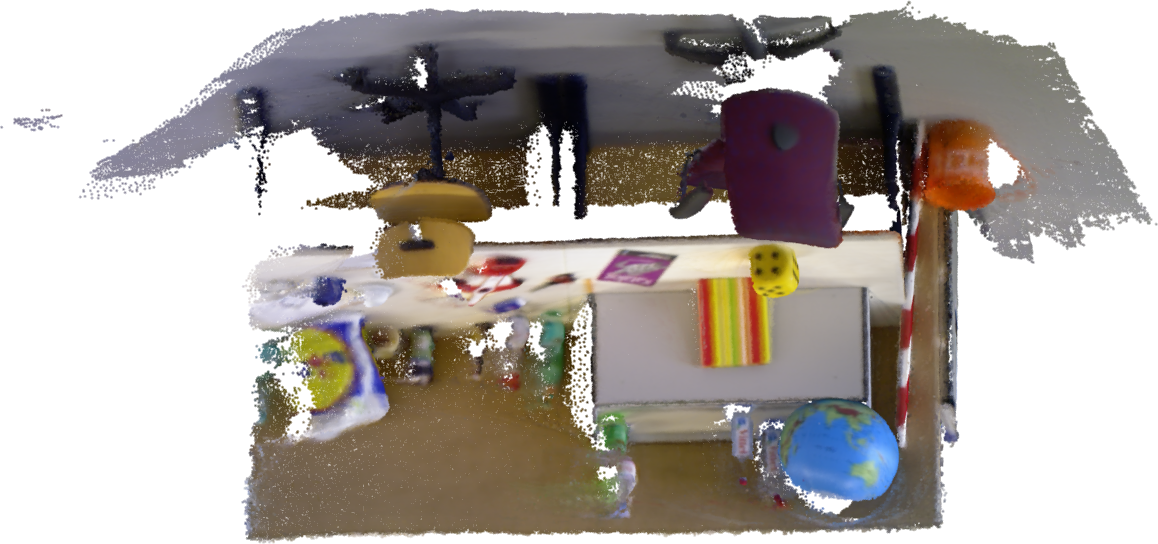}
    \hfill
    \includegraphics[width=0.25\linewidth, angle=180]{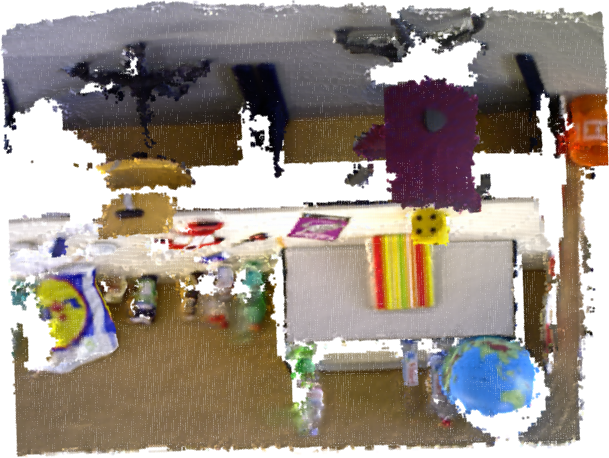} 
\vspace{1.5mm}
\caption{The reconstruction results on TUM RGB-D~\cite{sturm2012benchmark} sequences compare against two other camera tracking and refinement methods. While different methods perform better across sequences, our method achieves comparable results to the best-performed method. In addition, the texture is sharper and more precise.} \label{fg:tum_rgbd_results}
\end{figure}

We evaluate our method on the TUM RGB-D datasets to validate the camera pose refinement. Notably we take lighting conditions into account. The qualitative Root Mean Square Error (RMSE) of the absolute trajectory error is shown in Table~\ref{tab:tracking_results} of the main paper. Here we visualize in the Figure~\ref{fg:tum_rgbd_results} the refined point cloud of more sequences and compare with two other baseline methods: badslam~\cite{schops2019bad} and gradient-SDF~\cite{Sommer2022gradient}. For each sequence, we take $300$ frames as input to initialize SDF with $2$cm voxel size and the camera poses, then $30$ keyframes are chosen using a sharpness detector~\cite{bansal2016} to avoid manual selection. An up-sampling is applied after $10$ iterations.  

\section{Surface Geometry Refinement}
\subsection{Synthetic Data}
\begin{figure}[h]
\centering \includegraphics[width=0.18\linewidth]{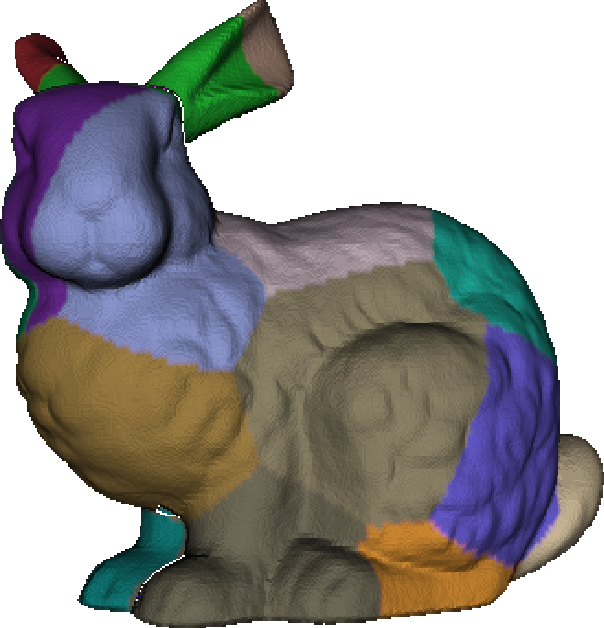}
\hfill
\centering\includegraphics[width=0.22\linewidth]{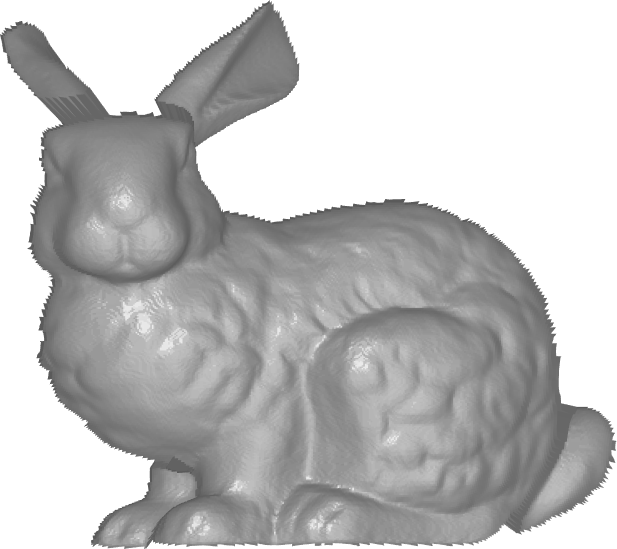}
\hfill
\centering\includegraphics[width=0.22\linewidth]{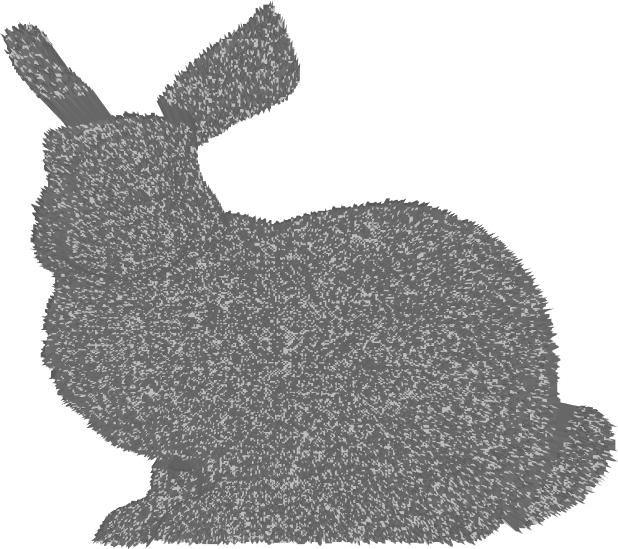}
\hfill
\centering\includegraphics[width=0.18\linewidth]{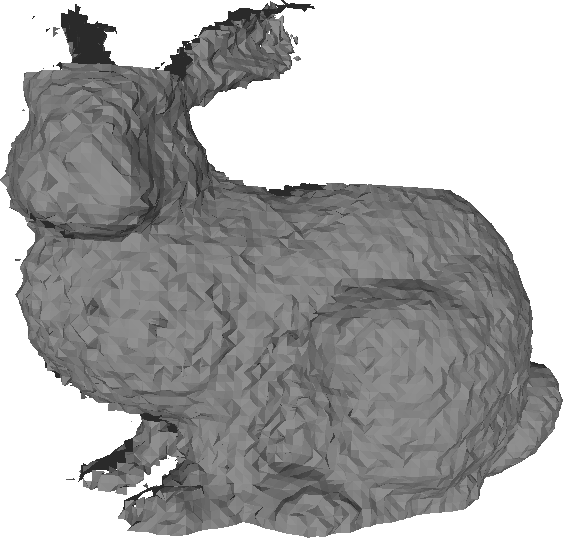}
\\
\centering\makebox[0.22\linewidth][c]{\small{RGB image}}
\hfill
\centering\makebox[0.22\linewidth][c]{\small{ground truth depth}}
\hfill
\centering\makebox[0.22\linewidth][c]{\small{noisy depth}}
\hfill
\centering\makebox[0.22\linewidth][c]{\small{initial mesh}} \\

\centering\includegraphics[width=0.18\linewidth]{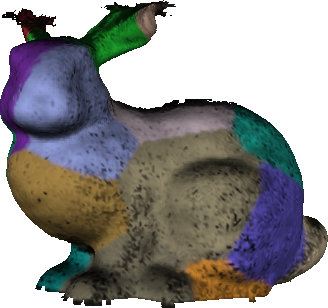}
\hfill
\centering\centering\includegraphics[width=0.18\linewidth]{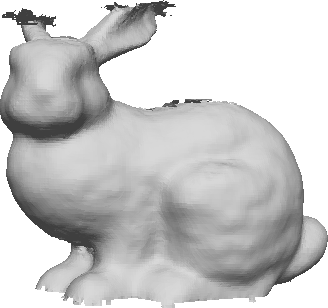}
\hfill
\centering\centering\includegraphics[width=0.18\linewidth]{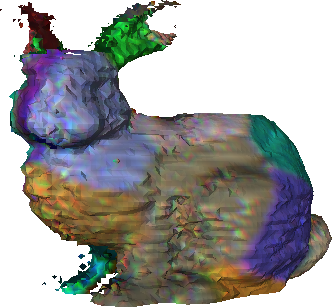}
\hfill
\centering\includegraphics[width=0.18\linewidth]{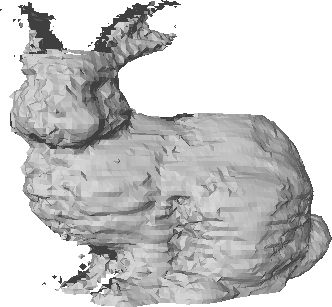} 
\\
\centering\makebox[0.22\linewidth][c]{\small{Intrinsic3d~\cite{maier2017intrinsic3d} colored mesh}}
\hfill
\centering\makebox[0.22\linewidth][c]{\small{Intrinsic3d~\cite{maier2017intrinsic3d} geometry}} 
\hfill
\centering\makebox[0.22\linewidth][c]{\small{\cite{bylow2019combining} albedo}}
\hfill
\centering\makebox[0.22\linewidth][c]{\small{\cite{bylow2019combining} geometry}}
\\
\centering\includegraphics[width=0.18\linewidth]{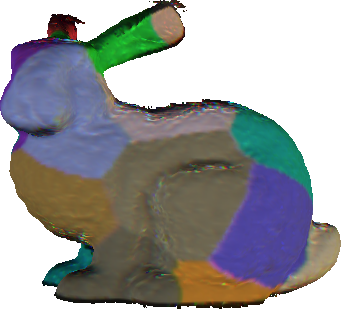}
\hfill
\centering\includegraphics[width=0.18\linewidth]{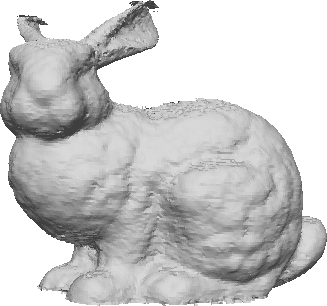}
\hfill
\centering\includegraphics[width=0.18\linewidth]{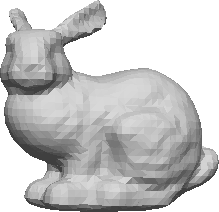} 
\hfill
\centering\includegraphics[width=0.18\linewidth]{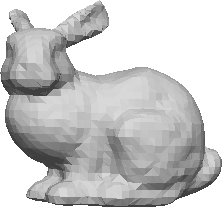}
 \\
 \centering\makebox[0.22\linewidth][c]{\small{Ours albedo}}
\hfill
\centering\makebox[0.22\linewidth][c]{\small{Ours geometry}}
\hfill
\centering\makebox[0.22\linewidth][c]{\small{volSDF~\cite{Yariv2021} geometry}}
\hfill
\centering\makebox[0.22\linewidth][c]{\small{NeuS~\cite{wang2021neus} geometry}}
\vspace*{1.5mm}
\caption{The rendered synthetic dataset bunny~\cite{turk1994zippered} with Kinect-like noise~\cite{khoshelham2012accuracy}. The initial mesh is the fused mesh after the camera tracking stage. The instrinsic3d~\cite{maier2017intrinsic3d} is over smoothed, and the mesh of~\cite{bylow2019combining} fails to deal with noisy depth. The work~\cite{Yariv2021} and~\cite{wang2021neus} does not recover albedo and the details are still missing.}\label{fg:bunny}
\end{figure}

We compare our results with two classical methods~\cite{bylow2019combining, maier2017intrinsic3d} and two neural-rendering-based methods~\cite{wang2021neus,Yariv2021}. Figure~\ref{fg:bunny} shows the input RGB image with ground truth depth and noisy depth we use for validate methods. Our mesh accomplishes more detailed surface reconstruction and a more faithful albedo. Note that the work of Maier el at. ~\cite{maier2017intrinsic3d} does not recover albedo but only recovers the gradient of its albedo. The color on the mesh is the average intensities of each voxel. The work~\cite{bylow2019combining} estimates the albedo but fails to recover it with good quality. With only RGB and camera poses as input, volSDF~\cite{Yariv2021} and NeuS~\cite{wang2021neus} can recover the mesh without color with a satisfactory quality but both works need train over 24 hours, and the resolutions are still not good. 

\subsection{Real-World Datasets}
\begin{figure}[h] 
    \centering
        \includegraphics[width=0.22\linewidth]{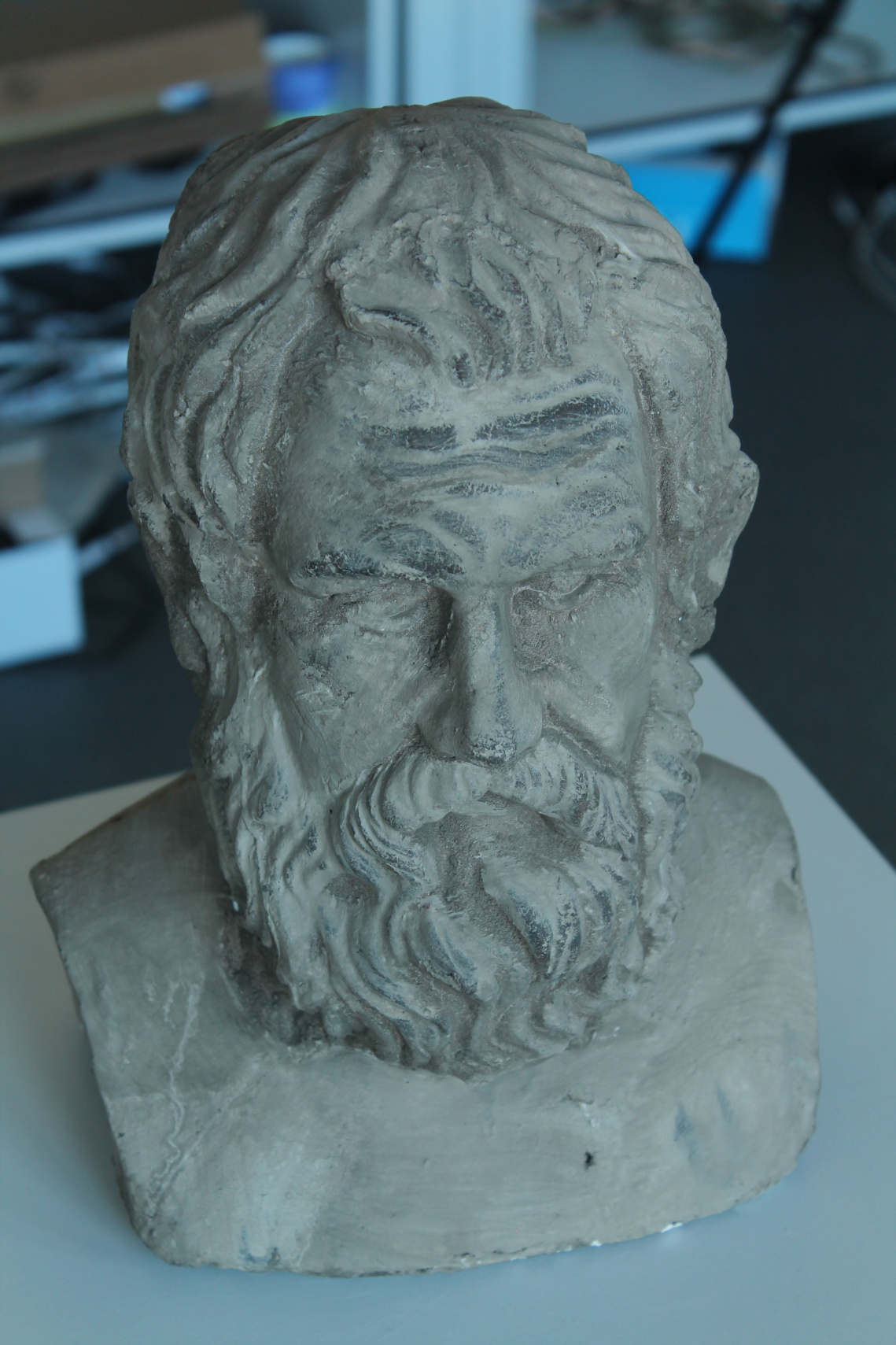} 
        \hfill
        \centering \includegraphics[width=0.20\linewidth, scale=0.3]{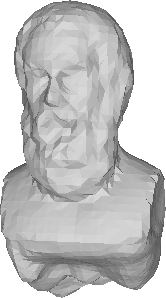} 
        \hfill
        \centering \includegraphics[width=0.22\linewidth]{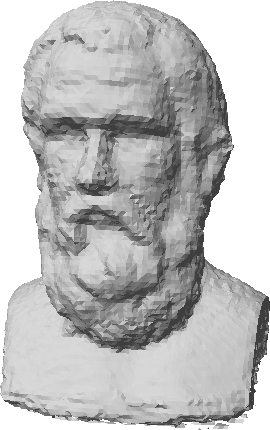} 
        \hfill
        \centering \includegraphics[width=0.22\linewidth, scale =1.1]{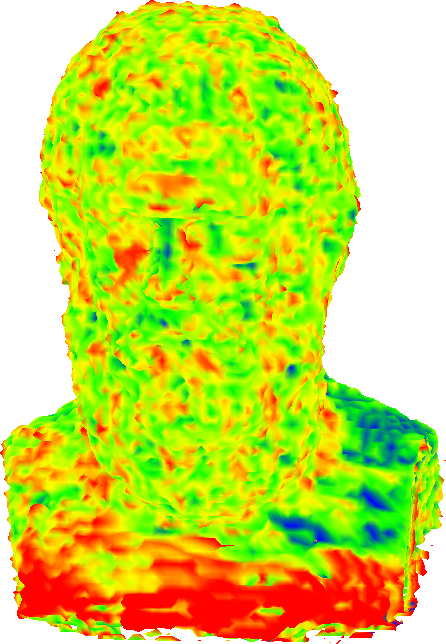} \\
        \vspace*{1.5mm}
        \makebox[0.22\linewidth][c]{\small{RGB image}}
        \hfill
        \makebox[0.22\linewidth][c]{\small{volSDF~\cite{Yariv2021}}}
        \hfill
        \makebox[0.22\linewidth][c]{\small{~\cite{bylow2019combining}}}
        \hfill
        \makebox[0.22\linewidth][c]{\small{~\cite{bylow2019combining} error}} \\
        \centering \includegraphics[width=0.22\linewidth]{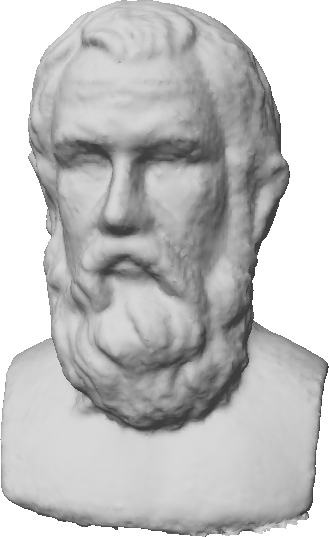}  
        \hfill
        \centering \includegraphics[width=0.22\linewidth]{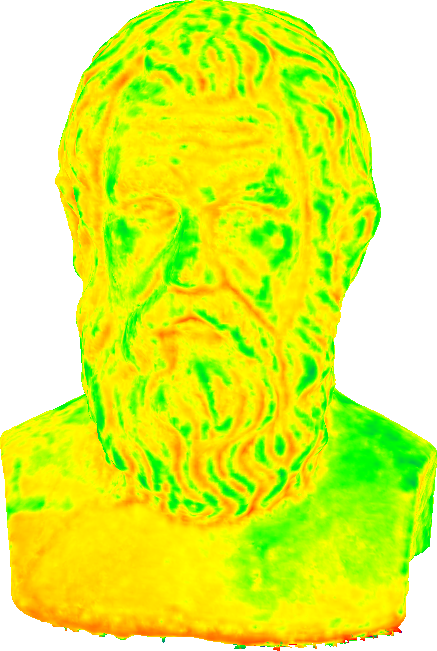}
        \hfill
        \centering \includegraphics[width=0.22\linewidth]{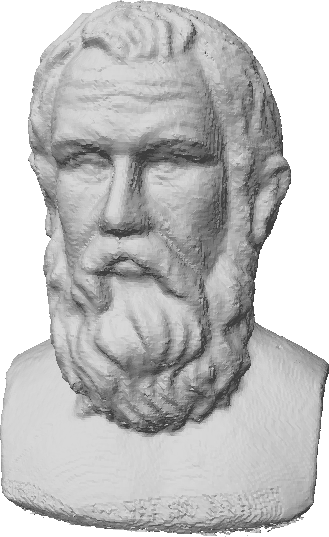}  
        \hfill
        \centering \includegraphics[width=0.22\linewidth]{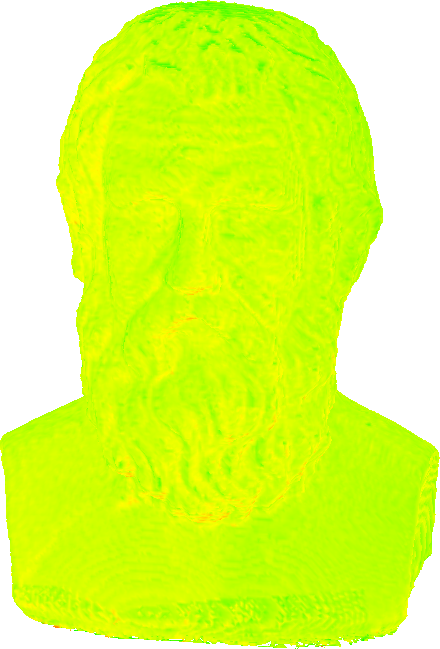}
        \\ \vspace{0.15cm}
        \makebox[0.19\linewidth][c]{\small{Intrinsic3d~\cite{maier2017intrinsic3d}}}
        \hfill
        \makebox[0.19\linewidth][c]{\small{Intrinsic3d~\cite{maier2017intrinsic3d} error}}
        \hfill
        \makebox[0.19\linewidth][c]{\small{Ours}}
        \hfill
        \makebox[0.19\linewidth][c]{\small{Ours error}}
    \vspace{1.5mm}
    \caption{Comparison results with the ground truth laser model of multi-view dataset Sokrates~\cite{zollhofer2015shading}. While NeuS~\cite{wang2021neus} fails to give any mesh, volSDF~\cite{Yariv2021} gives only coarse estimated surface, the proposed method achieves error (standard deviation) of $1.2$mm and Intrinsic3D~\cite{maier2017intrinsic3d} and~\cite{bylow2019combining} have error $2.1$mm and $3.7$mm respectively.}
    \label{fg:compare2}
\end{figure}

To measure the refined surface accuracy of our method quantitatively, we compare our results with a ground truth laser scan of the Socrates and Vase Multi-View Stereo dataset~\cite{zollhofer2015shading}, these datasets offer the color images, and corresponding depth images, together with the pre-estimated camera poses. The depth images have been masked. All the outside areas are set to zero. In the error map, the green color indicates the minor error, from the yellow to red color transaction indicates a larger positive distance to the ground truth, and the blue direction indicates the negative distance to the ground truth. Figure~\ref{fg:compare2} shows the comparison results of the related methods with our method. For the neural-rendering-based methods, volSDF~\cite{Yariv2021} can only recover the blurry shape while the NeuS~\cite{wang2021neus} fails to give any reconstruction surface, even though masks are provided. For classical methods, the work~\cite{bylow2019combining} suffers from the noisy depth, while intrinsic3d~\cite{maier2017intrinsic3d} can recover a good 3D model. However, the proposed method can recover the surface's fine scaled detail and achieve the smallest standard deviation error.

\section{Visualization of Multi-view and Recorded Dataset}
\begin{figure}[ht]
\begin{tabular}{c c}
   initial mesh & refined mesh \\
\includegraphics[width=0.3\linewidth]{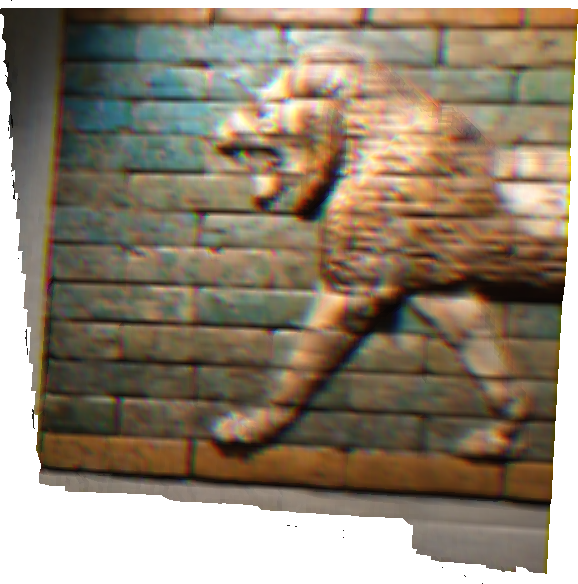}
&
\includegraphics[width=0.3\linewidth]{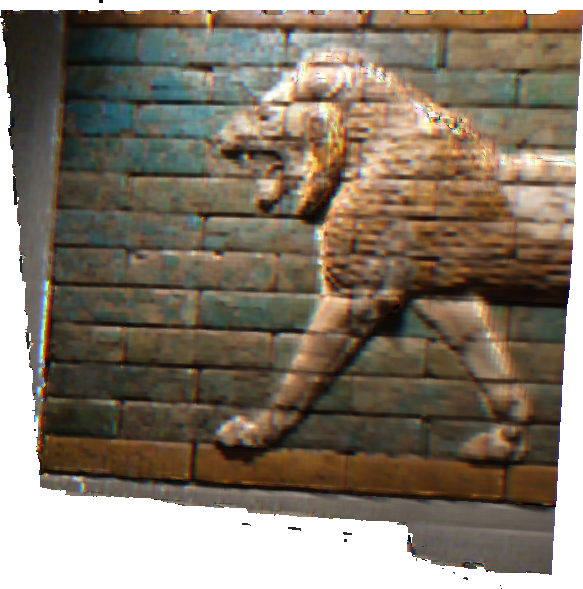}
\\
\includegraphics[width=0.3\linewidth, angle=180, origin=c]{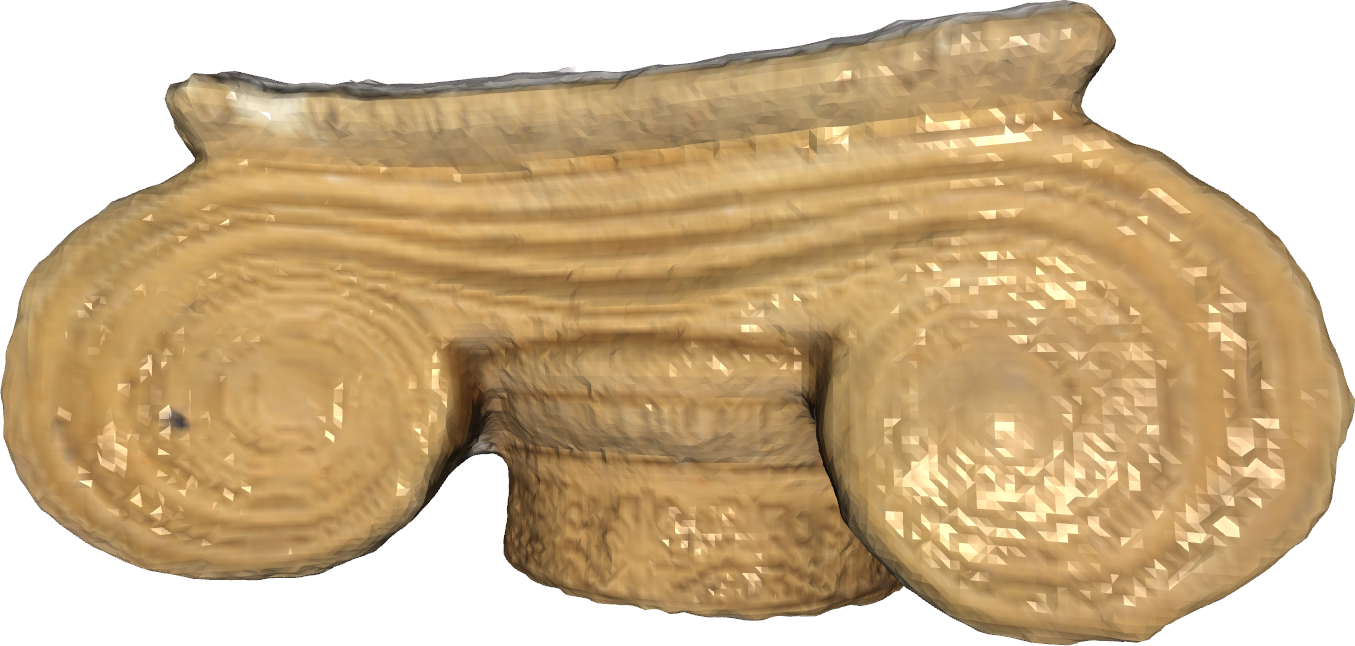}
&
\includegraphics[width=0.3\linewidth, angle=180, origin=c]{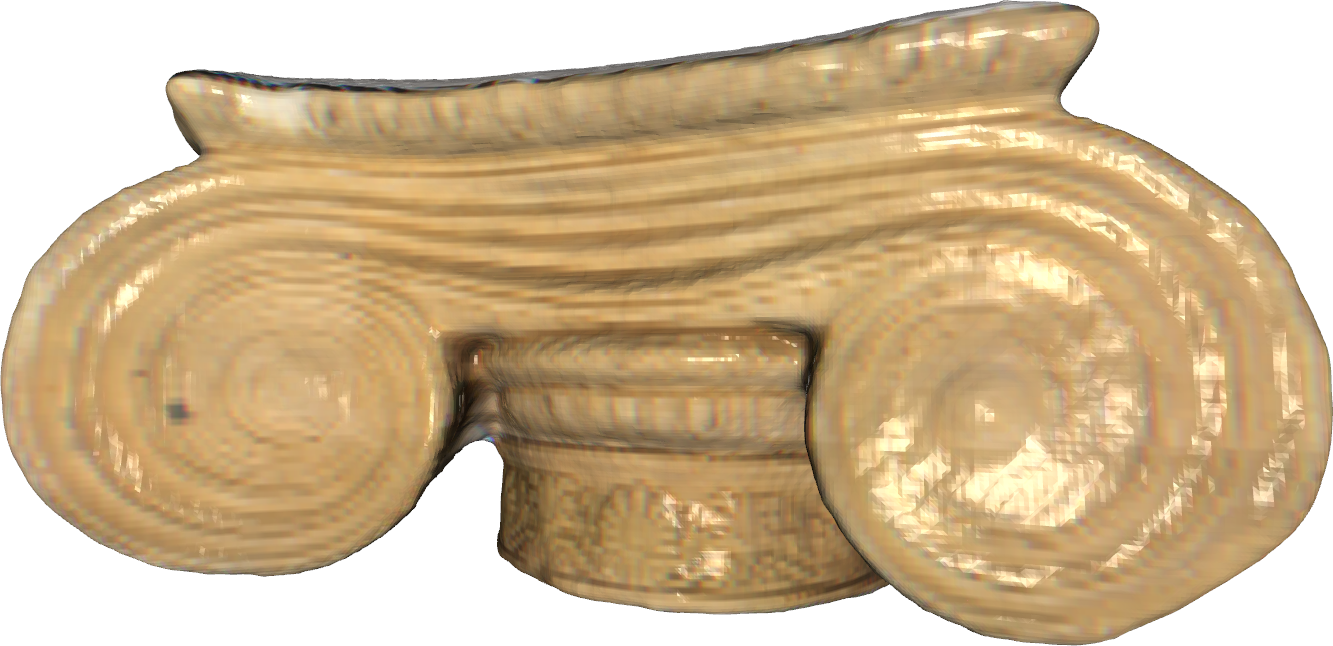}
\\
\includegraphics[width=0.45\linewidth]{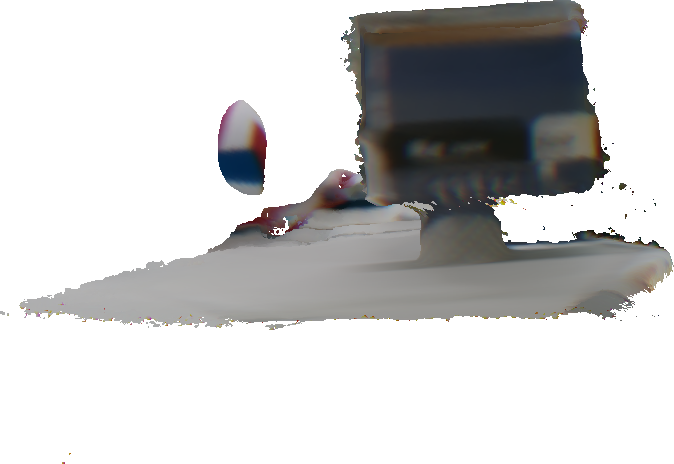}
&
\includegraphics[width=0.45\linewidth]{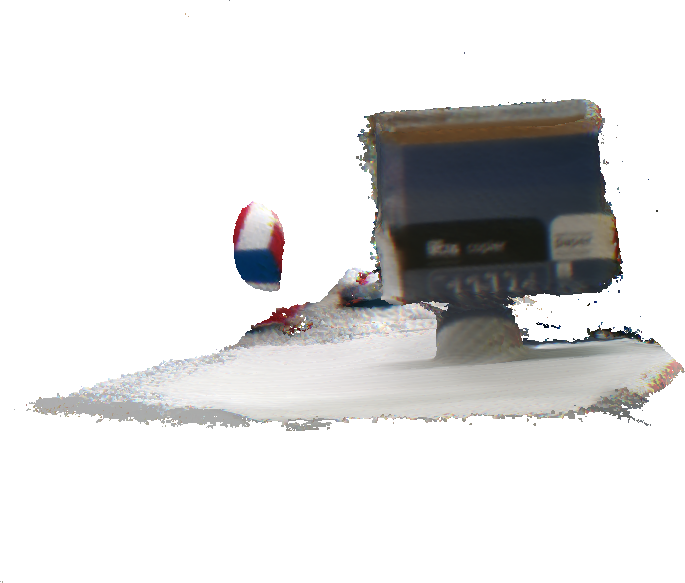}
\\
 \includegraphics[width=0.4\linewidth]{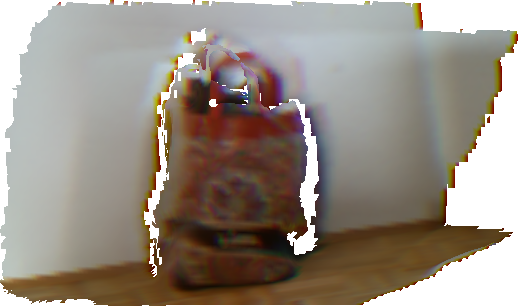}
&
\includegraphics[width=0.4\linewidth]{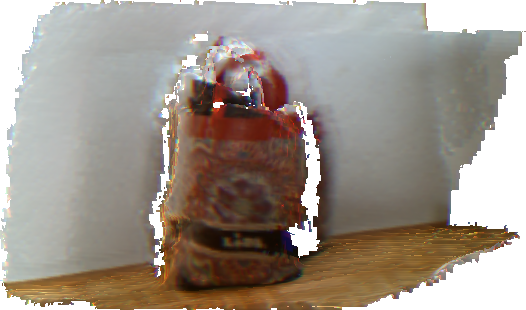}
\end{tabular}
\caption{The first two rows are the results of Lion dataset~\cite{maier2017intrinsic3d} and figure dataset~\cite{zollhofer2015shading}, we track $200$ frames and take $20$ keyframes. After the refinement, the texture is more clear and the geometry detail is recovered. The last two rows are the recorded datasets. The refined mesh recovered clear recognizable pattern and the original color of the object.}\label{fg:real-world}
\end{figure}

In this section, we demonstrate on Figure~\ref{fg:real-world} more visualization results of the 3D scene dataset~\cite{zhou2014color} and our own recorded dataset using the setup shown in Figure~\ref{fg:set_up} in the paper. Both datasets have no camera poses offered. 
\newpage

\section{Related Mathematics}
\paragraph*{List of Mathematical Symbols}
Here we list the mathematical symbols we used in the paper for a reference. 
\begin{table}[ht]
    \centering
    \begin{tabular}{|c |c| c| c|}
        \hline
        Symbol & Description & Symbol & Description \\
       \hline
       \hline
        $\vt{I}_i$ & $i$-th color image & $\vt{l}^s$ & vector from point light source location to point $\vt{x}$    \\ \hline
        $\vt{p}$ & continuous 2D image point in $\mathbb{R}^2$ & $\mathbb{S}^2$ & 2D sphere  \\ \hline
        $\vt{x}$ & continuous point in $\mathbb{R}^3$ & $\mat{R}_i$ & rotation matrix of frame $i$ \\ \hline
        $\vt{p}(\vt{x})$ & image points projected by space point $\vt{x}$ & $\vt{t}_i$ & translation vector of frame $i$ \\ \hline
        $\bmr(\vt{x})$ & reflectance (albedo) of point $\vt{x}$ & $\vt{o}$ & origin of camera coordinates \\ \hline
        $\vt{L}(\vt{i},\vt{x})$ & point $\vt{x}$'s incoming radiance of direction $\vt{i}$  & $d_{\surf{S}}(\vt{x})$ & minimum distance from point $\vt{x}$ to the surface $\surf{S}$  \\ \hline
        $\vt{v}^j$ & position of voxel $j$ in $\mathbb{R}^3$ & $\psi^j$ & voxel $j$ distance to the closest surface point \\ \hline
        $\vt{g}^j$ & voxel $j$ gradient & $\vt{n}(\vt{x})$ & surface normal at point $\vt{x}$ \\ \hline
        $\max(\cdot,\cdot)$ & max operator, take the larger one & $\mu_s$ & anisotropy coefficient of point light source $s$ \\ \hline
        $\SH(\cdot)$ & spherical harmonics function & $\nabla$ &gradient operator \\ \hline
        $\Psi^s$ & point light source light intensity &  $l_i$  & lighting coefficient in SH model at $i$-th frame \\ \hline
        $ \mathcal{V}$ & volume of the voxel grid & $\mathcal{I}$ & input image set \\ \hline
        $\nu_i^j$ & visibility indicator of voxel $j$ at image $i$ & $\M{}{\cdot, \cdot}$ & image formation model \\ 
        \hline
    \end{tabular}
    \vspace*{5mm}
    \caption{Summary of used mathematical symbols.}
\end{table}



\section{Mathematical Details of Optimization}
\subsection{Image Formation Model}
In this section, we illustrate the optimization details of the proposed method. The energy function is as described in equation~\eqref{eq:energy}. The optimization with robust estimator $\Phi$ is performed using the re-weighted least square method. In this section, we explain two proposed models in detail.\par

\paragraph*{SH model} From~\eqref{eq:5} and~\eqref{eq::SH_model_multi_view}, note that $\vt{x}_i$ is the point $\vt{x}$ under $i$-th image coordinates. We use the fact that first order SH model is linear, so $\dotp{\vt{l}_i, \SH(\mat{R}_i^\top\vt{x}^j)} = \dotp{\mat{R}_i\vt{l}_i, \SH(\vt{n}(\vt{x}^j))}$.  The total energy for SH model is 
\begin{align}
\min_{\{\mat{R}_i, \vt{t}_i, \vt{l}_i\}_i, \{\bmr^j, \psi^j\}_j} \vt{E}(\bmr^j, \psi^j, \mat{R}_i, \vt{t}_i, \vt{l}_i) &= \sum_{i,j}\nu_i^j\Phi(\vt{I}_i(\pi(\mat{R}_i^\top(\vt{x}^j-\vt{t}_i))) - \bmr^j\dotp{\mat{R}_i\vt{l}_i, \SH(\vt{n}(\vt{x}^j))})\nonumber \\
&+ |\norm{\nabla \psi^j} -1|^2 \,.
\end{align}
We directly optimize for $\hat{\vt{l}}_i = \mat{R}_i\vt{l}_i$ instead in the real experiment.
\paragraph*{PLS model} From~\eqref{eq::LED_model_multi_view}, the total energy for PLS model is 
\begin{align}
    \min_{\{\mat{R}_i, \vt{t}_i, \Psi^s_i\}_i, \{\bmr^j, \psi^j\}_j} &\vt{E}(\bmr^j, \psi^j, \mat{R}_i, \vt{t}_i, \Psi^s_i) \nonumber \\
    &= \sum_{i,j}\nu_i^j\Phi(\vt{I}_i(\pi(\mat{R}_i^\top(\vt{x}^j-\vt{t}_i))) - \Psi^s_i \bmr^j\frac{\max(\dotp{\mat{R}_i^\top\vt{n}(\vt{x}^j),-\mat{R}_i^\top(\vt{x}^j-\vt{t}_i)}, 0)}{\norm{\mat{R}_i^\top(\vt{x}^j-\vt{t}_i)}^3}) \nonumber \\
    &+ |\norm{\nabla \psi^j} -1|^2 \,.
\end{align}
\paragraph*{Optimization}
 Let $\vt{r}_i^j = \vt{I}_i^j - \bmr^j\M{}{\vt{x}^j, \X_i}$ and $\vt{r}_e^j = \norm{\nabla \psi^j} -1$. For the $k$th iteration, we optimize 
\begin{equation}
    \min \sum_{i,j} w(\vt{r}_i^{j,(k)}) (\vt{r}_{i}^j)^2 + (\vt{r}_e^j)^2  = 0\,, \label{eq:residual}
\end{equation}
where the weight function is defined as $w(\vt{r}) = \frac{\Phi(\vt{r})}{\vt{r}}$, and computed using current residuals.
\paragraph{Camera pose} The camera pose is updated using $6$DoF presentation. For $\omega_i \in SE(3)$, the rotation matrix $\mat{R}_i$ is updated by 
\begin{equation}
    \mat{R}^{(k+1)} = \mat{R}^{(k)}\exp(-\omega_i)\,.
\end{equation}
Written~\eqref{eq:residual} in vector form, we solve the linear system
\begin{equation} 
    (\vt{J}_c^\top W \vt{J}_c + \lambda_d I)\exp(\omega_i) = \vt{J}_c^\top W \vt{J}_c
\end{equation}
to solve the increment of the rotation matrix, $\vt{J}_c$ is the Jacobian matrix of $\mat{R}_i$, and $\lambda_d$ is the damping parameters and $I$ is the diagonal of the $\vt{J}_c^\top W \vt{J}_c $. For SH model, $\vt{J}_c$ calculated as
\begin{equation}
    \vt{J}_c = \frac{d \vt{r}}{d\omega_i} = [\nabla_x I, \nabla_y I]\left.\frac{d\pi}{d\vt{q}}\right\vert_{\vt{q}=\mat{R}_i^\top(\vt{v}^j-\vt{g}_j\psi_j-\vt{t}_i)}\left[\mat{R}_i^\top(\vt{v}^j-\vt{g}^j\psi^j-\vt{t}_i)\right]_{\times}\,,
\end{equation}
and for near field light is 
\begin{equation}
    \vt{J}_c =\frac{d \vt{r}}{d\omega_i} = [\nabla_x I, \nabla_y I]\left.\frac{d\pi}{d\vt{q}}\right\vert_{\vt{q}=\mat{R}_i^\top(\vt{v}^j-\vt{g}^j\psi^j-\vt{t}_i)}\left[\mat{R}_i^\top(\vt{v}^j-\vt{g}^j\psi^j-\vt{t}_i)\right]_{\times} - \left[\bmr^j\M{\vt{v}^j, \mat{R}_i, \vt{t}_i}\right]_{\times}\,.
\end{equation}
Here $\pi$ is the projection operator and $(\cdot)_{\times}$ is the twist matrix of the vector. \par
The translation vector is updated by $\vt{t}_i^{(k+1)} = \vt{t}^{(k)}_i - \Delta \vt{t}_i$, $\nabla \vt{t}_i$ is solved using the same step as $\exp(\omega_i)$. The Jacobian of $\vt{t}_i$ for LED model is 
\begin{equation}
    \frac{d \vt{r}}{d \vt{t}_i} = [\nabla_x I, \nabla_y I]\left.\frac{d\pi}{d\vt{q}}\right\vert_{\vt{q}=\mat{R}_i^\top(\vt{v}^j-\vt{g}^j\psi^j-\vt{t}_i)}\mat{R}_i^\top\,,
\end{equation}
and for the near field light the Jacobian is calculated using lagged attenuation term, that is we fix $\norm{(\vt{x}_{i}^j)^{(k)}}^3$ from last iteration and treat it as a constant, then 
\begin{equation}
    \frac{d \vt{r}}{d \vt{t}_i} = [\nabla_x I, \nabla_y I]\left.\frac{d\pi}{d\vt{q}}\right\vert_{\vt{q}=\mat{R}_i^\top(\vt{v}_j-\vt{g}^j\psi^j-\vt{t}_i)}\mat{R}_i^\top - \frac{\bmr^j}{\norm{\vt{x}_{i}^{j(k)}}^3} \vt{g}_i^\top \,.
\end{equation}

\paragraph{Voxel Distance} As mentioned in the paper, the surface normal $\vt{n}(\vt{x}^j)$ is approximated by normalized voxel gradient $\vt{g}^j$, thus it is the function of voxel distance $\psi^j$. In practice, the derivative of distance is calculated using (forward or backward) finite difference, the $\psi^j_x$, $\psi^j_y$, $\psi^j_z$ is the neighbor voxel in $x$, $y$, $z$ direction respectively.
\begin{equation}
    \vt{n}(\vt{x}^j) = \frac{\nabla \vt{g}^j}{\norm{\nabla \vt{g}^j}} = \frac{(\psi^j - \psi^j_{x}, \psi^j - \psi^j_{y}, \psi^j - \psi^j_{z} )}{\norm{(\psi^j - \psi^j_{x}, \psi^j - \psi^j_{y}, \psi^j - \psi^j_{z} )}}\,.
\end{equation}
Then for the natural light model, the Jacobian of $\psi^j$ is 
\begin{equation}
    \vt{J}_{\psi^j} = [\nabla_x I, \nabla_y I]\left.\frac{d\pi}{d\vt{q}}\right\vert_{\vt{q}=\mat{R}_i^\top(\vt{v}^j-\vt{g}^j\psi^j-\vt{t}_i)} \mat{R}_i^\top(\frac{d \vt{g}^j}{d \psi}\psi -\vt{g}^j) - \bmr^j\vt{l}_i^\top \frac{d \SH(\vt{g}^j)}{d \vt{g}^j}\frac{d \vt{g}^j}{d \psi^j}\,,
\end{equation}
and for the near field light model, for simplicity, like for camera translation update, we fix the attenuation term from last update, and use the fact that $\norm{\vt{g}^j} \approx 1$, then the Jacobian $\vt{j}_{\psi^j}$ is 
\begin{equation}
    \vt{J}_{\psi^j} = [\nabla_x I, \nabla_y I]\left.\frac{d\pi}{d\vt{q}}\right\vert_{\vt{q}=\mat{R}_i^\top(\vt{v}^j-\vt{g}^j\psi^j-\vt{t}_i)} \mat{R}_i^\top(\frac{d \vt{g}^j}{d \psi^j}\psi^j -\vt{g}^j) + \frac{\bmr^j}{\norm{\vt{x}_{i}^{j,(k)}}^3}\left[(\vt{v}^j - 2 \psi^j \vt{g} -\vt{t}_i)^\top\frac{d \vt{g}^j}{d \psi^j} - 1 \right]\,,
\end{equation}
Then the distance increment $\nabla\psi^j$ is also solved by solving $\vt{J}_{\psi^j}^\top W \vt{J}_{\psi^j} \nabla \psi^j = \vt{J}_{\psi^j}^\top W \vt{r} $. 
\paragraph{Reflectance and lighting} For $\bmr^j$ and lighting $\vt{l}_i$, the Jacobian matrices are also computed by simple $ \frac{d \vt{r}}{d \bmr^j}$, and $ \frac{d \vt{r}}{d \vt{l}_i}$, they are relative straight forward to compute compare to distance and camera rotations, so we omit the detail here. 

\subsection{Camera Tracking}
We use the SDF tracking energy to optimize the camera pose~\cite{Sommer2022gradient, bylow2013real}. To find the rigid body motion $\mat{R}$ and $\vt{t}$ for incoming point cloud with points $\vt{x}^k$ to the global shape $\surf{S}$. The energy is 
\begin{equation}
    \label{eq:tracking}
    \min_{\mat{R}, \vt{t}} E(\mat{R},\vt{t}) = \sum_{k}{w^kd_\surf{S}(\mat{R}\vt{x}^k+\vt{t})^2}\,,
 \end{equation}
note that $k$ is point index in the current depth. For $w^k = \max(\min(1+\frac{d^k}{T}, 1), 0)$ and 
\begin{equation}
    |d_\surf{S}(\vt{x})| = \min_{\vt{x}^s \in \surf{S}}\norm{\vt{x}-\vt{x}^s}\,.
\end{equation}
In our case
\begin{align}
&d_\surf{S}(\vt{x}) = \psi^{j*} + (\vt{x} - \vt{v}^{j*})^\top \vt{g}^{j*} \,,\\
&\nabla d_\surf{S}(\vt{x}) = \vt{g}^{j*} \,, \\
& j* = \argmin_j\norm{\vt{x} - \vt{v}^{j}}\,.
\end{align}
After the $R$, $t$ is optimized, the distance for current global shape $\surf{S}$ can be updated using current point cloud by weighted average from starting frame until the current frame 
\begin{align}
&\psi =\sum_i \frac{w_id_i}{w_i}\, \\
&d_i = (\vt{x}^{k*} - \mat{R}_i^\top(\vt{v}-\vt{t}_i))_z\, \\
&k* = \argmin_k \norm{\vt{x}^k - \mat{R}_i^\top(\vt{v}-\vt{t}_i)}\,.
\end{align}

\end{document}